\documentclass[10pt,twocolumn,letterpaper]{article}

\usepackage{cvpr}              %

\usepackage{graphicx}
\usepackage{amsmath}
\usepackage{amssymb}
\usepackage{booktabs}
\usepackage{enumitem}
\usepackage{xcolor}
\usepackage{placeins}
\usepackage[accsupp]{axessibility}

\usepackage[pagebackref,breaklinks,colorlinks]{hyperref}

\usepackage[capitalize]{cleveref}
\crefname{section}{Sec.}{Secs.}
\Crefname{section}{Section}{Sections}
\Crefname{table}{Table}{Tables}
\crefname{table}{Tab.}{Tabs.}

\definecolor{naturalcolor}{rgb}{0.004, 0.451, 0.698}
\definecolor{specializedcolor}{rgb}{0.871, 0.561, 0.02}
\definecolor{structuredcolor}{rgb}{0.008, 0.62, 0.451}
\newcommand{\naturalsym}{{\protect\scalebox{1.5}{\color{naturalcolor!100}$\bullet$}}}
\newcommand{\specializedsym}{{\protect\scalebox{1.5}{\color{specializedcolor!100}$\bullet$}}}
\newcommand{\structuredsym}{{\protect\scalebox{1.5}{\color{structuredcolor!100}$\bullet$}}}

\def \cS {\mathcal S}

\def \cD {\mathcal D}

\def \cM {\mathcal M}

\def\cvprPaperID{6400} %
\def\confName{CVPR}
\def\confYear{2022}

\begin{document}

\title{Which Model to Transfer? Finding the Needle in the Growing Haystack}

\author{Cedric Renggli\thanks{Work done while interning at Google Research. Correspondence to C. Renggli (\href{mailto:cedric.renggli@inf.ethz.ch}{cedric.renggli@inf.ethz.ch}) and M. Lucic (\href{mailto:lucic@google.com}{lucic@google.com}).}\\
ETH Zurich
\and André Susano Pinto\\
Google Research
\and Luka Rimanic\\
ETH Zurich
\and Joan Puigcerver\\
Google Research
\and Carlos Riquelme\\
Google Research
\and Ce Zhang\\
ETH Zurich
\and Mario Lucic\\
Google Research
}

\maketitle

\begin{abstract}
    Transfer learning has been recently popularized as a data-efficient alternative to training models from scratch, in particular for computer vision tasks where it provides a remarkably solid baseline. The emergence of rich model repositories, such as TensorFlow Hub, enables the practitioners and researchers to unleash the potential of these models across a wide range of downstream tasks. As these repositories keep growing exponentially, efficiently selecting a good model for the task at hand becomes paramount. 
    We provide a formalization of this problem through a familiar notion of \emph{regret} and introduce the predominant strategies, namely \emph{task-agnostic} (e.g.\ ranking models by their ImageNet performance) and \emph{task-aware} search strategies (such as \emph{linear} or \emph{kNN} evaluation). We conduct a large-scale empirical study and show that both task-agnostic and task-aware methods can yield high regret. We then propose a simple and computationally efficient hybrid search strategy which outperforms the existing approaches. We highlight the practical benefits of the proposed solution on a set of 19 diverse vision tasks.
\end{abstract}

\vspace{-1em}
\section{Introduction}
\vspace{-0.5em}

Services such as TensorFlow Hub or PyTorch Hub\footnote{\url{https://tfhub.dev} and \url{https://pytorch.org/hub}}
offer a plethora of pre-trained models that often achieve state-of-the-art performance on specific tasks in the vision domain. The predominant approach, namely choosing a pre-trained model and \emph{fine-tuning} it to the downstream task, is an effective and data efficient approach~\cite{huh2016makes,kolesnikov2019revisiting, ngiam2018domain,raghu2019transfusion, yosinski2014transferable,zhai2019visual}. Perhaps surprisingly, this approach is also effective when the pre-training task is significantly different from the target task, such as when applying an ImageNet pre-trained model to diabetic retinopathy classification~\cite{oquab2014learning}.
Fine-tuning often entails adding several more layers to the pre-trained deep network and tuning all the parameters using a limited amount of downstream data. Due to the fact that all parameters  are being updated, this process can be extremely costly %
in terms of compute~\cite{zhai2019visual}. Fine-tuning all models to find the best performing one is becoming computationally infeasible. A more efficient alternative is to simply train a cheap
classifier on top of the learned representation (e.g.\ pre-logits). However, the performance gap with respect to fine-tuning can be rather large~\cite{kolesnikov2019revisiting, kornblith2019better}.

This raises a very practical question: \textit{Given a new task, how to pick the best model to fine-tune?} This question was intensively studied in recent years and existing approaches can be divided into two groups:
(a) \textit{task-agnostic} model search strategies, which rank pre-trained models independently of the downstream task (e.g.\ ranking the models by ImageNet accuracy, if available)~\cite{kornblith2019better}, and (b) \textit{task-aware} model search strategies, which make use of the downstream dataset in order to rank models (e.g.\ $k$NN classifier accuracy as a proxy for fine-tuning accuracy, or using the meta-learned Task2Vec representations)~\cite{achille2019task2vec,meiseles2020source,puigcerver2020experts}. Most of the prior work attempts to answer this question by using a homogeneous sets of pre-trained models whereby the models share the same architecture or they were trained on the same dataset. However, this does not reflect the current landscape of models available in online repositories. 

As a result, several practically relevant questions remained open. Firstly, how do existing methods compare in the presence of both ``generalist models'' (e.g.\ models trained on a relatively diverse distribution such as ImageNet) and ``expert models'' (e.g.\ trained on domain-specific datasets such as plants), across a diverse collection of datasets? Secondly, \textit{is there a method which strikes a good balance between computational cost and performance?}

\textbf{Our contributions.} 
In this paper, we provide a large-scale, systematic empirical study of these questions.
(i) We define and motivate the model search problem through a notion of regret. We conduct the first study of this problem in a realistic setting focusing on heterogeneous model pools. 
(ii)~We consider 19 downstream tasks on a heterogeneous set of 46 models grouped into 5 representative sets. (iii)~We highlight the dependence of the performance of each strategy on the constraints of the model pool, and show that, perhaps surprisingly, both task-aware and task-agnostic proxies suffer a large regret on a significant fraction of downstream tasks. (iv) Finally, we develop a hybrid approach which generalizes across model pools as a practical alternative.

\begin{figure*}[t]
\centering\captionsetup{width=1\linewidth}
\includegraphics[width=0.9\textwidth]{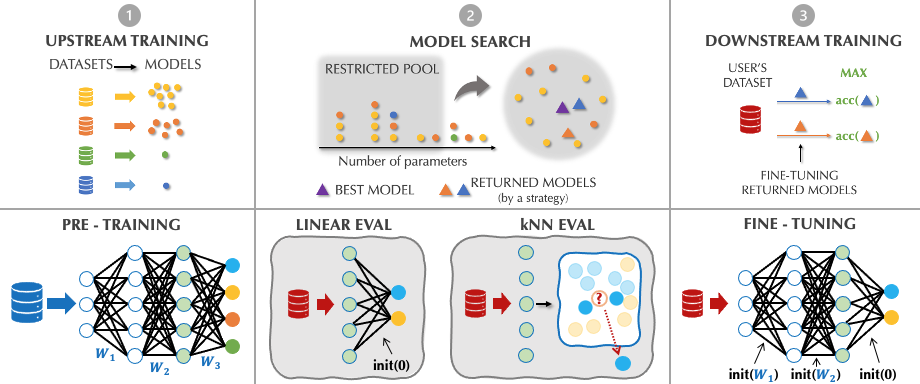}
\caption{Transfer learning setup: \textbf{(1) Upstream models} Pre-training of models from randomly initialized weights on the (large) upstream datasets; 
\textbf{(2) Model search} Either downstream task independent or by running a proxy task, i.e.\ fixing the weights of all but the last layer and training a \textit{linear classifier} or deploying a \textit{kNN classifier} on the downstream dataset; \textbf{(3) Downstream training} Unfreezing all the weights, optimizing
all layers (incl. pre-defined ones) on the downstream dataset.}
\label{fig:Illustration_all}
\vspace{-1em}
\end{figure*}

\vspace{-0.5em}
\section{Background and related work}\label{sec:background}
\vspace{-0.5em}

We will now introduce the main concepts behind the considered transfer learning approach where the pre-trained model is adapted to the target task by learning a mapping from the intermediate representation to the target labels~\cite{pan2009survey, tan2018survey, wang2018theoretical, weiss2016survey}, as  illustrated in Figure~\ref{fig:Illustration_all}.

\textbf{(I) Upstream models.} \quad The variety of data sources, losses, neural architectures, and other design decisions,  leads to a variety of \emph{upstream models}. The user has access to these models, but cannot control any of these dimensions, nor access the upstream training data.

\begin{figure}[t]
\centering\captionsetup{width=\linewidth}
\includegraphics[width=0.45\textwidth]{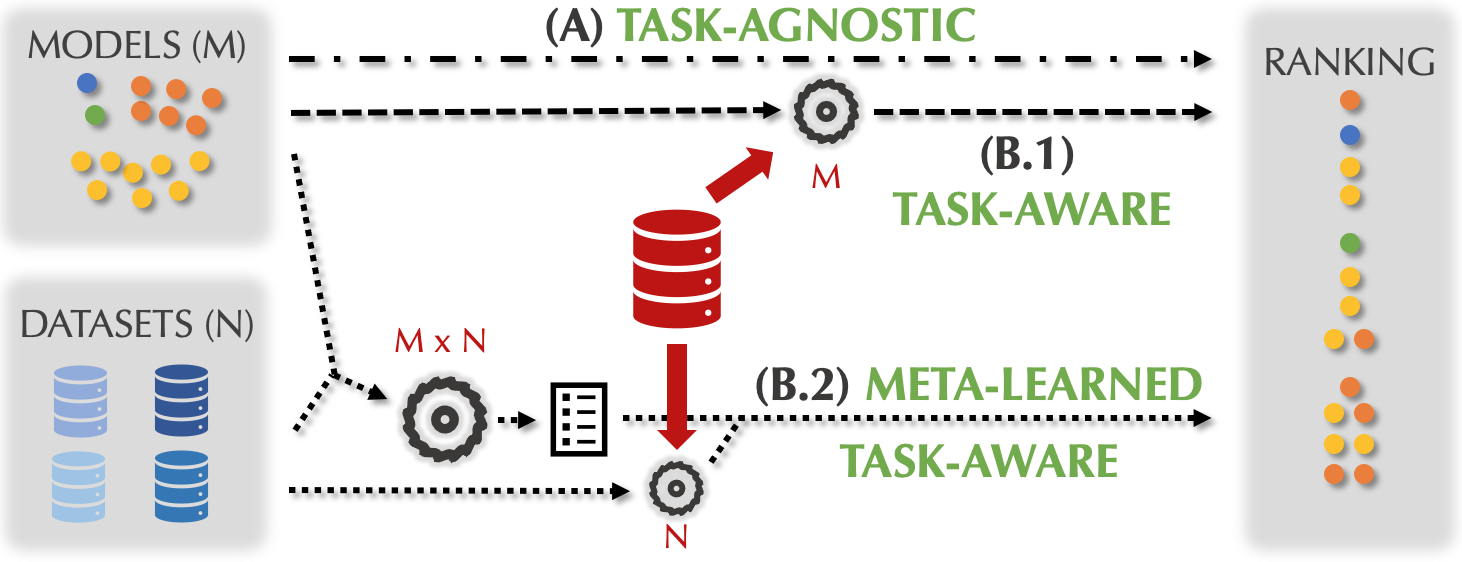}
\caption{Model search methods: \textbf{(A) Task-agnostic} methods do not see the downstream task, producing the same ranking of models for all possible tasks (e.g.\ using the highest ImageNet accuracy); \textbf{(B.1) Task-aware} methods deploy a proxy (e.g.\ linear evaluation) for each model on user's dataset
; \textbf{(B.2) Meta-learned task-aware} methods use a collection of datasets ahead of time for the subsequent model search (e.g.\ Task2Vec~\cite{achille2019task2vec}, exploring dataset similarities and copying the ranking of the closest dataset).}
\label{fig:model_search_methods}
\vspace{-1em}
\end{figure}

\textbf{(II) Model search.}\quad Given no limits on computation, the problem becomes trivial -- exhaustively fine-tune each model and pick the best performing one. In practice, however, one is often faced with stringent requirements on computation, and the aim of the second stage in Figure~\ref{fig:Illustration_all} is therefore to select a relatively small set of models for fine-tuning. Selecting this set of models is the central research question of this paper (cf.~Figure~\ref{fig:model_search_methods}).

\textbf{(II A) Task-agnostic search strategies.}\quad These approaches rank models before observing the downstream data~\cite{kornblith2019better} and consequently select the same model for \emph{every} task. The most popular approach can be summarized as follows (i) Pick the model with the highest test accuracy on ImageNet, otherwise (ii) pick the one trained on the largest dataset. If there is an (approximate) tie, pick the model with most parameters.

\textbf{(II B.1) Task-aware search strategies.}\quad In contrast to the task-agnostic approach, task-aware methods may use the downstream data, thus requiring additional computation. The idea is to extract the learned representations from the pretrained model, train a linear or a $k$NN classifier on those representations, and select the models that achieve highest accuracy. Note that this requires at least one forward-pass for each instance of the dataset, and for every model, so the computational complexity grows linearly with the model pool size. Nevertheless, this approach is usually several orders of magnitude faster than fine-tuning. Strictly speaking, one can apply an early stopping strategy to the fine-tuning procedure which allows a finer control of the accuracy-time tradeoff. We study this in Section~\ref{sec:exp_ablation}, whereas in Section~\ref{sec:other_related_work} we discuss and contrast with other task-aware search strategies.

\textbf{(II B.2) Meta-learned task-aware search strategies.}\quad 
The idea is to calculate the transfer learning performance for each task in a benchmark set. Given a new dataset, the model search part is performed by finding the best model(s) of the nearest benchmark task, for instance given by learned task embedding on all the benchmark tasks and the downstream dataset. While a single task embedding can be computed efficiently by training a single probe-network, meta-learned approaches require to initially evaluate the performance of all models across each benchmark task. In addition, the list of transfer learning performances needs to be expanded whenever either new models or new benchmark datasets are added. We have simulated this setting by considering the most prominent such approach, namely Task2Vec~\cite{achille2019task2vec}, and provide some preliminary results in Appendix~\ref{sec:appendix:task2vec}. In Section~\ref{sec:other_related_work}, we elaborate this approach, along with other related work in this category (e.g.\ such as semantic-based search strategies~\cite{dwivedi2020DDS, song2020depara, zamir2018taskonomy}), and contrast it to the search strategies we consider in this work.

\textbf{(III) Downstream training.}\quad
In this stage, the selected model is adapted to the downstream task (cf.~Figure~\ref{fig:Illustration_all}). The predominant approach is to fully or partially apply the pre-trained neural network as a feature extractor. The head (e.g.\ last linear layer) of the pre-trained model is replaced with a new one, and the whole model is trained on the target data. This process is commonly referred to as \textit{fine-tuning} and it often outperforms other methods~\cite{donahue2014decaf, kornblith2019better, oquab2014learning, sharif2014cnn}. 

\vspace{-0.5em}
\section{Computational budget and regret}\label{sec:regret}
\vspace{-0.5em}

The main aim of this work is the study of simple methods to filter and search pre-trained models before stepping into the --more expensive-- fine-tuning process. Formally, we define a search method $m(\cM, \cD)$ with budget $B$ as a function which takes a set of models $\cM$ and a downstream dataset $\cD$ as input, and outputs a set of distinct models $\cS_m \subseteq \cM$, with $\vert \cS_m \vert = B$. Those $B$ models are then all fine-tuned to obtain the best test accuracy on the downstream task $\cD$.

\textbf{Budget and regret.}\quad Fine-tuning represents the largest computational cost so, accordingly, we define the \emph{number} of models that are fine-tuned as the computational complexity of a given method. Given any fixed budget $B$, we would like to return a set $\cS$ which includes the models resulting in good performance downstream. In particular, we define the notion of \emph{absolute regret} of a search strategy $m$ and a pool of models $\cM$ on dataset $\cD$ as  \vspace{-1em}
\begin{equation} \vspace{-0.8em}
    \label{eq:regret}
    \underbrace{\max_{m_i \in \cM} \mathbf{E}[t(m_i, \cD)]}_{\textsc{oracle}} - \underbrace{\mathbf{E}\left[\max_{s_i \in \cS_m} t(s_i, \cD)\right]}_{s(m)},
\end{equation}
where $t(m, \cD)$ is the test accuracy achieved when fine-tuning model $m$ on dataset $\cD$.
The first expectation is taken over the randomness in the $t(\cdot)$ operator, that is, the randomness in fine-tuning 
and due to a finite sampled test set.
In addition to the randomness in $t(\cdot)$, the second expectation also accounts for any potential randomization in the algorithm that computes $\cS_m$. We define $s(m)$ as the expected maximal test accuracy achieved by any model in the set $\cS_m$, the set of models returned by a fixed strategy $m$.
In our case, $k$NN is deterministic as all the downstream data is used, whereas the linear model depends on the randomness of stochastic gradient descent.
To enable comparability between datasets of different difficulty as well as a comparison between two selection strategies $m_1$, and $m_2$, we define their \textit{relative delta} as \vspace{-0.4em}
\begin{equation}\vspace{-0.4em}
    \label{eq:rel_delta}
    \Delta(m_1, m_2) := \frac{s(m_1) - s(m_2)}{1-\min (s(m_1), s(m_2))}, 
\end{equation}
with $s(\cdot) \in [0, 1]$ as defined in Equation~\ref{eq:regret}. Substituting $s(m_1)$ by the $\textsc{oracle}$ value, and $s(m_2)$ by $s(m)$ leads to the \textit{relative regret} $r(m)$.
We discuss the impact of alternative notions in Section~\ref{sec:exp_ablation}.

\vspace{-0.5em}
\section{Experimental design}\label{sec:experimentalsetup}
\vspace{-0.5em}

Our goal is to assess which model search strategies achieve low regret when presented with a diverse set of models. As discussed, there are three key variables: (i) The set of downstream tasks, which serve as a proxy for computing the expected regret of any given strategy, (ii) the \emph{model pool}, namely the set we explore to find low-regret models, and (iii) the transfer-learning algorithms.

\vspace{-0.5em}
\subsection{Datasets and models}
\vspace{-0.5em}

\textbf{Datasets.}\quad
We use VTAB-1K, a few-shot learning benchmark composed of 19 tasks partitioned into 3 groups  -- \naturalsym\emph{natural}, \specializedsym\emph{specialized}, and \structuredsym\emph{structured}~\cite{zhai2019visual}. The \emph{natural} image tasks include images of the natural world captured through standard cameras, representing generic objects, fine-grained classes, or abstract concepts. \emph{Specialized} tasks contain images captured using specialist equipment, such as medical images or remote sensing. The \emph{structured} tasks are often derive from artificial environments that target understanding of specific changes between images, such as predicting the distance to an object in a 3D scene (e.g.\ DeepMind Lab), counting objects (e.g.\ CLEVR), or detecting orientation (e.g.\ dSprites for disentangled representations). Each task has $800$ training examples, $200$ validation examples, and the full test set. This allows us to evaluate model search strategies on a variety of tasks and in a setting where transfer learning offers clear benefits with respect to training from scratch~\cite{zhai2019visual}.

\newcommand{\jftslices}{\textsc{Expert}}
\newcommand{\imnetacc}{\textsc{ImNetAccuracies}}
\newcommand{\maxresnet}{\textsc{ResNet-50}}
\newcommand{\restrdim}{\textsc{Dim2048}}
\newcommand{\all}{\textsc{All}}

\textbf{Models.}\quad
The motivation behind the model pools is to simulate 
several use-cases that are ubiquitous in practice. We 
collect 46 classification models (cf. Appendix~\ref{app:ModelDetails}):
\begin{itemize}[leftmargin=7.5mm,topsep=0mm,itemsep=0pt, nolistsep]
    \item 15 models trained on the ILSVRC 2012 (ImageNet) classification task~\cite{russakovsky2015imagenet}, including Inception V1-V3 models~\cite{szegedy2016rethinking}, ResNet V1 and V2 (depth 50, 101, and 152)~\cite{he2016deep}, MobileNet V1 and V2~\cite{howard2017mobilenets}, NasNet~\cite{zoph2018learning} and PNasNet~\cite{liu2018progressive} networks.
    \item 16 ResNet-50-V2 models trained on (subsets of) JFT~\cite{puigcerver2020experts}. These models are trained on different subsets of a larger dataset and perform significantly better on a small subset of downstream tasks we consider (i.e.\ they can be considered as \emph{experts}).
    \item 15 models from the VTAB benchmark\footnote{\url{https://tfhub.dev/vtab}}, with a diverse coverage of losses (e.g.\ generative, self-supervised, self-supervised combined with supervised, etc.) and architectures. In all cases the upstream dataset was ImageNet, but the evaluation was performed across the VTAB benchmark which does not include ImageNet.
\end{itemize}

\vspace{-0.5em}
\subsection{Model pools}
\label{sec:model-pools}
\vspace{-0.5em}

\begin{figure*}[!t]
\centering\captionsetup{width=1.0\linewidth}
\includegraphics[width=0.9\textwidth]{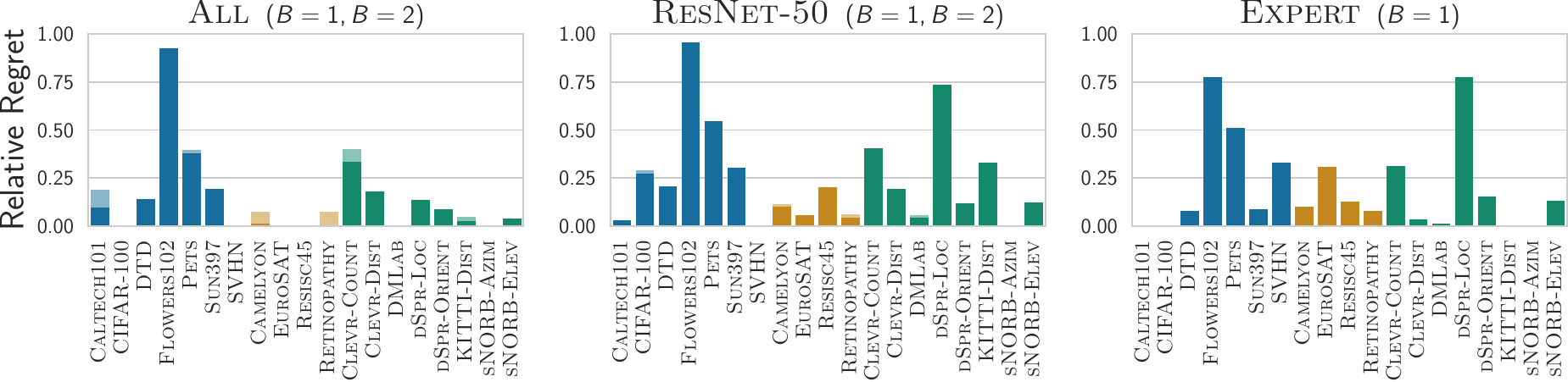}
\caption{\textbf{Task-agnostic strategies.} Relative regret ($r(m)$, cf. Section~\ref{sec:regret}) with $B=1$ (transparent) and $B=2$ (solid) on the \all, \maxresnet\, and $\jftslices$ pools, bearing in mind that there is only one task-agnostic model in $\jftslices$. By definition, task-agnostic strategies exclude experts yielding high regret on the \maxresnet\, and \jftslices\ pools, particularly on natural or structured datasets.}
\label{fig:task_agnostic_fails_relative}
\vspace{-1em}
\end{figure*}

\textbf{(A) Identifying good resource-constrained models (\maxresnet, \restrdim).} \quad
Here we consider two cases: (i) \maxresnet: All models with the number of parameters smaller or equal to ResNet50-V2. While the number of parameters is clearly not the ideal predictor, this set roughly captures the models with limited memory footprint and inference time typically used in practice. Most notably, this pool excludes the NasNet and PNasNet architectures, and includes the \emph{expert} models. (ii) \restrdim: The transfer strategies discussed in Section~\ref{sec:background} might be sensitive to the size of the representation. In addition, restricting the representation size is a common constraint in practical settings. Hence, we consider a model pool where the representation dimension is limited to a maximum of 2048.

\textbf{(B) Identifying expert models in presence of non-experts (\jftslices).}\quad
We consider a pool of 16 ResNet-50-V2 models from~\cite{puigcerver2020experts}. These models which we considered as \emph{experts} are trained on different subsets of a larger dataset. As the number of available models and the upstream training regimes increase, the number of such experts is likely to increase. As such, this presents a realistic scenario in which an expert for the target task may be present, but it is hard to identify it due to the presence of other models, some of which might perform well \emph{on average}.

\textbf{(C) Do better task-agnostic models transfer better (\imnetacc)?}\quad
This pool offers the ability to choose an upstream representation-learning technique that is best suited for a specific downstream task. This pool is mainly used to validate the idea that (a) ImageNet models transfer well across different tasks~\cite{he2019rethinking, huh2016makes} and that (b) better ImageNet models transfer better~\cite{kornblith2019better}.

\textbf{(D) All models (\all).}\quad
Finally, we consider the hardest setting, namely when the model pool contains all 46 models and no conceptual nor computational restrictions are in place. We note that: $\jftslices \subset \maxresnet \subset  \restrdim \subset \all$ and $\imnetacc \subset \all$.

\vspace{-0.5em}
\subsection{Evaluation procedures}\label{sec:eval_procedure}
\vspace{-0.5em}

\textbf{Fine tuning.}\quad
To assign a downstream test accuracy to each pair of model and task, we use the median test performance of 5 models obtained as follows: (i) Add a linear layer followed by a softmax layer and train a model on all examples of the training set. (ii) Fine-tune the obtained model twice, considering two learning rates, and 2500 steps of SGD and a batch size of 512~\cite{zhai2019visual}. (iii) Return the model with the highest validation accuracy. Note that in this case, the entire model, and not only the linear layer, is retrained. As a result, there are 10 runs for each model and we obtain 8740 trained models ($46 \times 19 \times 5 \times 2$).

\textbf{Linear evaluation.}\quad We train a logistic regression classifier added to the model representations (fixed) using SGD. We consider two learning rates ($0.1$ and $0.01$) for 2500 steps and select the model with the best validation accuracy. For robustness we run this procedure 5 times and take the median validation accuracy out of those resulting values. As a result, we obtain again 8740 models.

\textbf{$k$NN evaluation.}\quad We compute the validation accuracy by assigning to each of the 200 validation samples the label of the nearest training example (i.e.\ $k=1$) using standard Euclidean distance. 

\vspace{-0.5em}
\section{Key experimental results}
\vspace{-0.5em}

In this section we challenge common assumptions and highlight the most important findings of this study, whilst the extended analysis containing all the plots and tables can be found in the supplementary material. We remark that in the main body we only consider three main pools of models -- \all\,, \maxresnet\, and \jftslices, as we see them as the most representative ones. Since \restrdim\, behaves very similarly to \maxresnet, whereas \imnetacc\, is used only to confirm the findings of~\cite{kornblith2019better}, the results of ablation studies involving these two pools can be found in Appendix~\ref{sec:app_other_pools}. Finally, in this section we mainly investigate linear evaluation as the task-aware choice; all the corresponding plots for $k$NN can be found in Appendix~\ref{sec:app_full_task_aware}.

\begin{figure*}[!tb]
\centering\captionsetup{width=1.0\linewidth}
\includegraphics[width=0.9\textwidth]{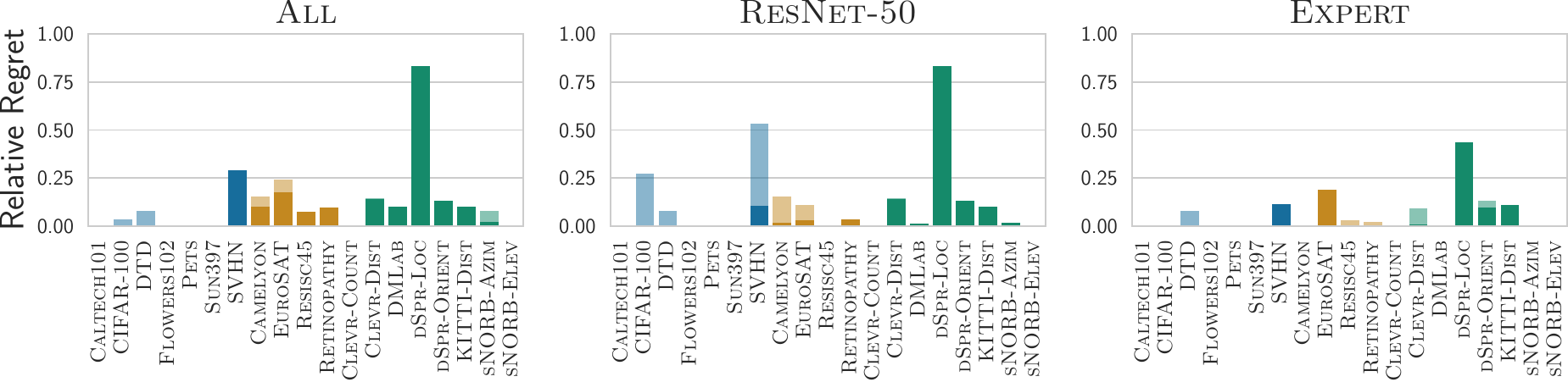}
\caption{\textbf{Task-aware strategies (linear).} Relative regret for $B=1$ (transparent) and $B=2$ (solid) on the \all, \maxresnet, and $\jftslices$ pools. Compared to task-agnostic strategies, we observe improvement on natural datasets (except SVHN) and on restricted pools (except $\textsc{dSpr-Loc}$), due to its ability to properly choose experts.}
\label{fig:linear_fails_relative}
\vspace{-1em}
\end{figure*}

\vspace{-0.5em}
\subsection{High regret of task-agnostic strategies}
\label{sec:task_agnostic}
\vspace{-0.5em}

Figure~\ref{fig:task_agnostic_fails_relative} shows the results for task-agnostic methods with budget $B=1$ and $B=2$ on the \all, \maxresnet, and $\jftslices$ pools.
We observe a significant regret, particularly for $\maxresnet$ and $\jftslices$ pools (30\% of the datasets have a relative regret larger than 25\% on those two pools). This highlights the fact that task-agnostic methods are not able to pick expert models, in particular on natural and structured datasets. As more experts become available, this gap is likely to grow, making it clear that task-agnostic strategies are inadequate on its own. 

\vspace{-0.5em}
\subsection{Are task-aware always a good predictor?}
\vspace{-0.5em}

Intuitively, having access to the downstream dataset should be beneficial. We evaluate both the linear and the $k$NN predictor as detailed in Section~\ref{sec:experimentalsetup}. Figure~\ref{fig:linear_fails_relative} provides our overall results for the linear model, whereas analogous results for $k$NN are presented in Appendix~\ref{sec:app_full_task_aware}.
The method struggles on some structured datasets (in particular on $\textsc{dSpr-Loc}$). 

\begin{figure*}[!t]
\centering\captionsetup{width=1.0\linewidth}
\includegraphics[width=0.9\textwidth]{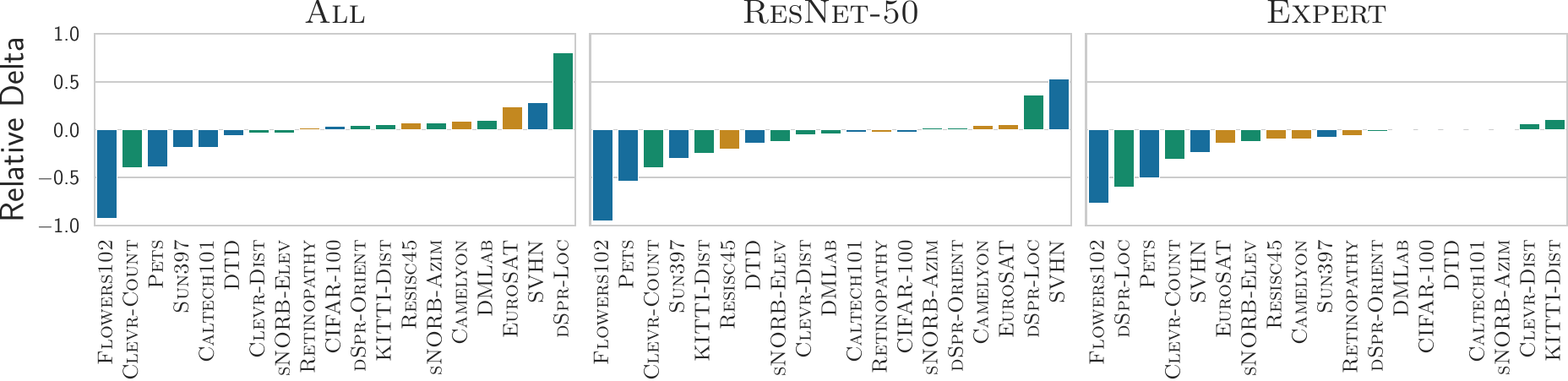}
\caption{\textbf{Task-agnostic} (positive if better) \textbf{vs Task-aware (linear)} (negative if better) for $B=1$. On the \all\, pool, the methods perform in a similar fashion, with respect to the number of datasets and the amount in which one outperforms the other. When one restricts the pool to \maxresnet\, or \jftslices\, task-aware methods outperform the task-agnostic method on most datasets. The relative delta is defined in Equation~\ref{eq:rel_delta} in Section~\ref{sec:regret}.}
\label{fig:task_aware_vs_task_agnostic_relative}
\vspace{-1em}
\end{figure*}

Compared to the task-agnostic strategy, as presented in Figure~\ref{fig:task_aware_vs_task_agnostic_relative} for $B=1$, we observe significant improvements on restricted model pools.
The \jftslices\, pool benefits the most: linear evaluation outperforms task-agnostic methods on almost every dataset (task-aware is only outperformed on three datasets by more than 1\%, and by 10\% in the worst case on the $\textsc{KITTI-Dist}$ dataset). On the other hand, task-agnostic and task-aware strategies seem to outperform each other on a similar number of datasets and by a comparable magnitude in the \all\, pool.
This suggests that no single strategy uniformly dominates all the others across pools.

In order to understand this further, we perform an ablation study where we plot the linear and $k$NN regret on the \imnetacc\, pool in Appendix~\ref{sec:app_other_pools}. In Figures~\ref{fig:linear_fails_other_pools_relative} and \ref{fig:knn_fails_other_pools_relative} we observe that task-aware search methods perform rather poorly when having access \emph{only} to different architectures trained on the same upstream data.
The \imnetacc\ models are included in the \all\, pool, and in some datasets some of those models are the best-performing ones.

Performance of the $k$NN predictor is on par on half of the datasets across the pools, and slightly worse than linear evaluation on the other half. We present these findings in Figure~\ref{fig:linear_vs_knn_relative} in Appendix~\ref{sec:app_full_task_aware}.

\begin{figure*}[!tb]
\centering\captionsetup{width=1.0\linewidth}
\includegraphics[width=0.9\textwidth]{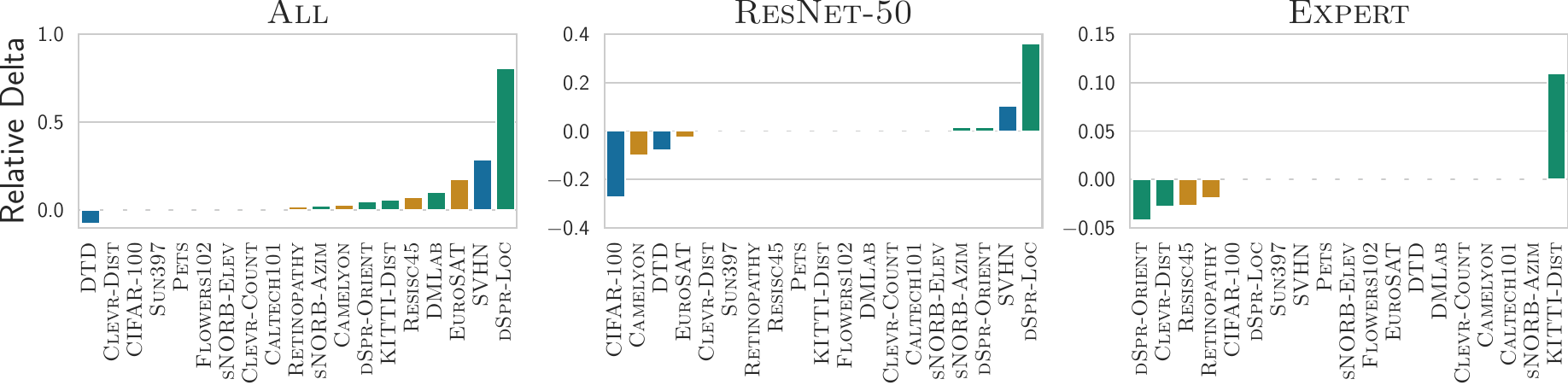}
\caption{\textbf{Hybrid linear} (positive if better) \textbf{vs Linear evaluation} (negative if better) for $B=2$. We observe that hybrid linear significantly outperforms linear with the same budget on the \all\, pool. Even though for \maxresnet\, and \jftslices\, pools there are datasets on which linear performs better than hybrid, the amounts in which it does are usually small. We note that most significant gains of hybrid come on certain structured datasets, the hardest task for every strategy.}
\label{fig:hybrid_vs_linear_relative}
\vspace{-1em}
\end{figure*}

\vspace{-0.5em}
\subsection{Hybrid approach yields the best of both worlds}
\vspace{-0.5em}

A hybrid approach that selects both the top-1 task-agnostic model and the top-$(B-1)$ task-aware models leads to strong overall results. Figure~\ref{fig:hybrid_vs_linear_relative} shows how the hybrid approach with linear evaluation as the task-aware method significantly outperforms its linear counterpart with $B=2$. 
This is most noticeable in the \all\, pool where the task-agnostic model provides a large boost on some datasets. 

As we saw in Figure~\ref{fig:task_aware_vs_task_agnostic_relative}, when looking at the \all\, pool, the task-agnostic candidate tends to beat the linear one on datasets such as $\textsc{dSpr-Loc}, \textsc{SVHN}$ or $\textsc{EuroSAT}$.
Similarly, the linear candidate model clearly outperforms its task-agnostic counterpart on many natural datasets such as $\textsc{Flowers}$ or $\textsc{Pets}$. A comparison of Figures~\ref{fig:task_aware_vs_task_agnostic_relative} and~\ref{fig:hybrid_vs_linear_relative} reflects how the dominance of linear-only strategy vanishes on most datasets when confronted with the hybrid approach.
For the \maxresnet\, and \jftslices\, pools, as expected, the hybrid algorithm preserves the good picks of the linear proxy.
That said, we observe an increase of 36\% on $\textsc{dSpr-Loc}$ in the \maxresnet, and 11\% on $\textsc{KITTI-Dist}$. Both are structured datasets on which the linear proxy task performs poorly, as shown in Figure~\ref{fig:linear_fails_relative}.

The hybrid strategy requires to fine-tune at least two models.
Given that it performs well across all model pools and datasets, this is a reasonable price to pay in practice, and we suggest its use as the off-the-shelf approach.
Figures~\ref{fig:hybrid_vs_knn_relative} and~\ref{fig:hybrid_linear_vs_hybrid_knn_relative} in Appendix~\ref{sec:app_full_task_aware} depict the results for $k$NN.
In \jftslices\ models, the second $k$NN pick tends to beat the task-agnostic one -- hurting the $k$NN hybrid outcomes.
Overall, the hybrid linear approach consistently outperforms the one based on $k$NN.

\vspace{-0.5em}
\subsection{Further ablation studies}
\vspace{-0.5em}
\label{sec:exp_ablation}

\begin{figure*}[!tb]
\centering\captionsetup{width=1.0\linewidth}
\includegraphics[width=0.9\textwidth]{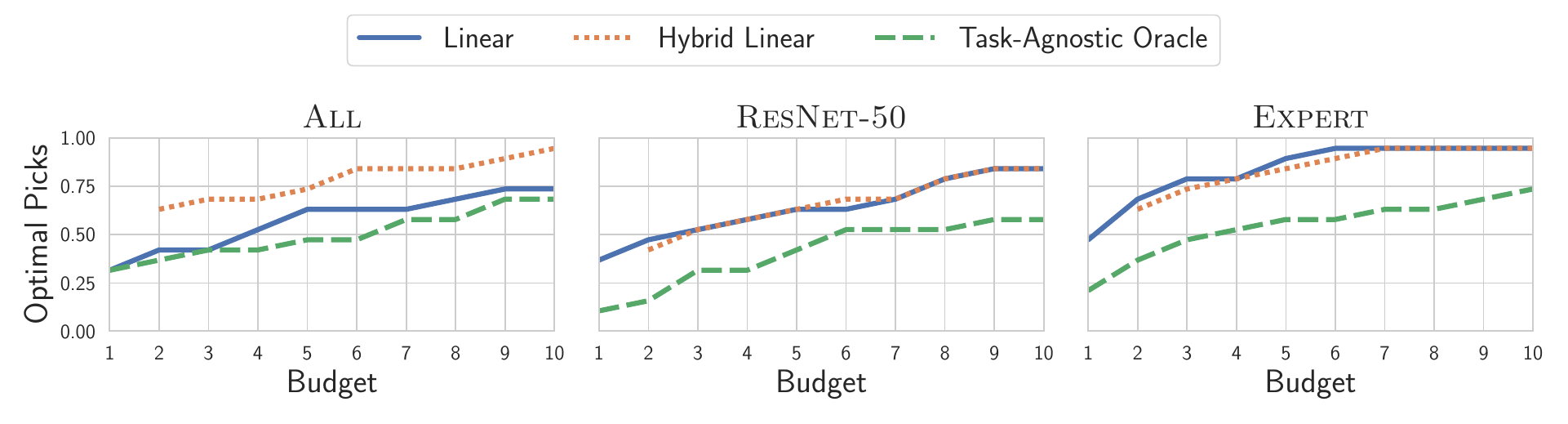}
\vspace{-1em}
\caption{\textbf{Optimal picks as a function of the computational budget.} The number of picked models (relative) with zero regret across three representative pools. We note that hybrid linear outperforms all other methods on \all, whilst being comparable with the linear strategy on restricted pools where linear alone already performs well. Here, the task-agnostic oracle refers to a method which ranks models based on their average accuracy across all datasets (more details Section~\ref{sec:exp_ablation}).
}
\label{fig:budget_vs_correct_picks}
\vspace{-1em}
\end{figure*}

\textbf{How does the computational budget impact the findings?}\quad
We have seen that for a limited budget of $B=2$ the proposed hybrid method outperforms the other strategies.
A natural question that follows is: how do these methods perform as a function of the computational budget $B$?
In particular, for each budget $B$, we compute how frequently does a strategy pick the best model.
The results are shown in Figure~\ref{fig:budget_vs_correct_picks}.  We observe that the hybrid linear strategy outperforms all individual strategies on the \all\, pool. Furthermore, it also outperforms a strong impractical task-agnostic oracle which is allowed to rank the models by the average fine-tune accuracy over all datasets. Our hybrid strategy achieves an on par performance with the linear approach on pools on which linear performs well. When task-aware strategies perform badly (e.g.\ pools without expert models), hybrid linear is significantly stronger (as seen in  Figure~\ref{fig:budget_vs_correct_picks_other_pools} in Appendix~\ref{sec:app_other_pools}). These empirical results demonstrate that the hybrid strategy is a simple yet effective practical choice for a model search strategy.

\textbf{Alternative evaluation scores.}\quad
Both \cite{kornblith2019better} and \cite{meiseles2020source} compute the correlation between the ImageNet test accuracy and the average fine-tune accuracy across datasets. Although this provides a good task-agnostic evaluation method for the \emph{average performance} (cf. Figure 1, right, in \cite{kornblith2019better}), it can be significantly impacted by outliers that have poor correlations on a specific dataset (cf. Figure 2, middle row, in \cite{kornblith2019better}). In Appendix~\ref{app:limitationCorrelation} we highlight another limitation of using correlation scores in the context of model-search strategies across heterogeneous pools.
Nevertheless, we empirically validate that ranking the models based on their ImageNet test accuracy on the \imnetacc\ pool transfers well to our evaluation setting (cf. Figure~\ref{fig:task_agnostic_fails_other_pools} in the Appendix~\ref{sec:app_other_pools}).
Furthermore, we show that reporting the differences of logit-transformed accuracies (log-odds) leads to similar conclusions as ours (cf.  Appendix~\ref{app:logVSrel}).
We opt for the relative regret $r(m)$, defined in Section~\ref{sec:regret}, as it is more intuitive and contained in $[-1,1]$. 

\textbf{Impact of the $k$NN hyperparameters.}\quad
The $k$NN classifier suffers from the curse of dimensionality~\cite{snapp1991asymptotic}, which is why we study the impact of the dimension (i.e.\ the representation size) on the $k$NN evaluation.
We fix a dataset, and plot a model's $k$NN score versus its representation dimension.
In order to have a single point per dimension and avoid an over-representation of the expert models that are all of the same architecture, we choose the model with the best $k$NN accuracy.
By calculating the Pearson correlation coefficient between the dimension and the respective $k$NN scores, we observe a moderate anti-correlation ($R<-0.5$) for 3 datasets, a moderate correlation ($R>0.5$) for 3 other datasets, and either small or no correlation for the remaining 13 datasets. Based on this empirical evidence we conclude that there is no significant correlation between the $k$NN classifier accuracy and the dimension. We provide more details in Figure~{\ref{fig:knn_vs_dim_coorelation}} of Appendix~\ref{app:knn_vs_dim}.
Regarding $k$, our preliminary experiments with $k=3$ offered no significant advantages over $k=1$.

\textbf{Early stopping and fine-tuning.}
\emph{Early stopping} approach, namely fine-tuning the entire network only for a small number of iterations, allows one to explore the accuracy vs. time tradeoff. However, it is understood that results heavily depend on the neural architectures and hyper-parameters used during pre-training (e.g. models trained with Batch normalization necessitate a different strategy). In our experiments, summarized in Appendix~\ref{sec:appendix:early_stopping}, we do not exhibit any benefits which generalize beyond certain specific settings that would favor this strategy over the simple linear proxy (cf. Figures~\ref{fig:earlystop_vs_task_agnostic_relative}).

\vspace{-0.5em}
\section{Other related work}
\vspace{-0.5em}
\label{sec:other_related_work}

Given access to the meta-data such as the source datasets, one could compare the upstream and downstream datasets~\cite{bhattacharjee2020p2l}; blend the source and target data by reweighting the upstream data to reflect its similarity to the downstream task~\cite{ngiam2018domain}; or construct a joint dataset by identifying subsets of the upstream data that are well-aligned with the downstream data~\cite{ge2017borrowing}. These approaches are restricted to one model per upstream dataset, while also being less practical as they necessitate training a new model as well as access to upstream datasets, which might be unavailable due to proprietary or privacy concerns.

Additionally, best-arm identification bandits algorithms suggest the successive elimination of sub-optimal choices \cite{even2006action} derived by fine-tuning models for shorter time. This combines model selection and downstream training, noting that it suffers from the same limitations as the partial fine-tuning proxy approach described Section~\ref{sec:exp_ablation}. 

\textbf{Other task-aware strategies.}
Alternative approaches such as H-Score~\cite{bao2019information}, LEEP~\cite{nguyen2020leep} or NCA~\cite{tran2019transferability} replace the computation of the linear classifier's weights on top of the frozen weights by using a cheaper estimator for the classifier's accuracy. These estimators are derived using a pseudo label distribution by analyzing the classification output of the pre-trained models. All three approaches have a clear limitation in supporting only those cases in which a classification head for the original task is provided together with each pre-trained model. Furthermore, the theoretical guarantees are given with respect to linear classification models trained on top of the frozen representations only. 
As observed with LEEP, albeit on natural datasets only, fine-tuning achieves significantly better test accuracies than a linear classifier, which further invites for better understanding of using linear accuracy as a proxy for fine-tuning, noting that linear correlations are not necessarily transitive. 
Furthermore, training a single linear layer, or calculating the kNN accuracy on top of transformed representations, is typically much less computational demanding than running the inference itself, and both steps can easily be overlapped. 
Recently, \cite{meiseles2020source} introduced the Mean Silhouette Coefficient (MSC) to forecast a model's performance after fine-tuning on time series data. We omit this approach due to its provable relation to a linear classifier proxy, whilst being similar to $k$NN with respect to capturing non-linear relationships.
Finally, \cite{puigcerver2020experts} utilize $k$NN as a cheap proxy task for searching models in a set of experts with the same architecture without quantifying and evaluating the regret.
LFC/LGC~\cite{deshpande2021linearized} are comparable to the early stopping strategy, which instead of using the gradient at the initialization point to compare models, follows this gradient for a small number of iterations. We note that this related work evaluates its proposed strategy in a fairly homogeneous model pool, a setting in which many task-aware methods and the corresponding early-stopping strategies have less room to fail.

\textbf{Meta-learned strategies.}
Meta-learned search strategies aim at reducing the linear complexity of having to run inference and training a subsequent model with the downstream dataset on all pre-trained models for performing a model search.
Such approaches typically contain a meta-learning part that precedes the actual search component by making use of a set of benchmark datasets. These datasets do not have to be related to the datasets used to train models upstream. In fact, in our scenario, the upstream datasets are unknown or inaccessible to the model search method, thereby rendering the benchmark datasets independent from the pre-trained models.
The computational complexity of performing the search is moved to the meta-learning part by calculating the task embedding and fine-tuning all pre-trained models on a set of benchmark datasets, as described in Section~\ref{sec:background} for Task2Vec~\cite{achille2019task2vec}. This computationally intensive task could be re-used by subsequent model search queries with different downstream datasets, but would typically need to be maintained by the provider  of pre-trained model repositories. 
We simulate the performance of Task2Vec in our experimental setting in Appendix~\ref{sec:appendix:task2vec}. We observe that meta-learned task-aware strategies can be improved by task-agnostic search methods (cf. Figure~\ref{fig:task2vec_vs_task_agnostic_relative}) for the budget of $B=1$, but not necessarily for a larger budget. Investigating the impact of benchmark datasets and larger fine-tune budget for these methods is beyond the scope of this work and represents an interesting line of future work.

Another line of related work is Taskonomy~\cite{zamir2018taskonomy} which finds the best source dataset(s), or nearest benchmark task in the meta-learned context, by exhaustively exploring the space of all possibilities. While it achieves promising results, it is not directly comparable to our approach: (i) the input domain and data is assumed to remain constant and tasks are only different in their labels, (ii) the chosen architecture keeps the weights of the encoder frozen, and the decoder trained on top usually consists of multiple fully connected layers, opposed to our fine-tuning regime with a single linear layer on top. Improvements by \cite{dwivedi2020DDS} and \cite{song2020depara} make Taskonomy faster, but they still do not distinguish multiple models trained on the same dataset, nor bypass the constraints described previously. Finally, for generalization of this approach beyond classification tasks, Mensink et al.~\cite{mensink2021factors}  extensively studied transfer behaviour across different types of tasks. They suggest that a task-agnostic strategy of selecting the model fine-tune on the largest possible dataset will perform well, whereas if a task-aware strategy is able to find a model trained on a most similar source, if it exists, it should outperform the task-agnostic strategy.

\vspace{-0.5em}
\section{Conclusions, limitations, and future work}
\vspace{-0.5em}

Transfer learning offers a data-efficient solution to train models for a range of downstream tasks.
As we witness an increasing number of models becoming available in repositories such as TensorFlow Hub, finding the right pre-trained models for target tasks is getting harder.
Fine-tuning all of them is not an option. 
In practice, the computational budget is limited and efficient model search strategies become paramount.
We motivate and formalize the problem of efficient model search through a notion of \emph{regret}, and argue that regret is better suited to evaluate search algorithms than correlation-based metrics.
Empirical evaluation results for the predominant strategies, namely \emph{task-agnostic} and \emph{task-aware} search strategies, are presented across several scenarios, showing that both can sometimes yield high regret. In other words, for any individual method we study, there exists a pool of models on which the method fails.
Finally, we propose a simple and computationally efficient hybrid search strategy which consistently outperforms the existing approaches over 19 diverse vision tasks and across all the defined model pools.

\textbf{Limitations and future work.} \quad In order to further stress-test the generalization of analyzed strategies, the number of relevant model pools could be increased by incorporating more diverse upstream tasks, in particular neural architectures, losses, and datasets. This would potentially yield more expert models, making the task both more challenging and more important, further highlighting the advantages of an effective search strategy. Secondly, we observe that task-aware methods consistently perform poorly in specific cases, such as when we consider diverse architectures trained only on ImageNet. There is no obvious reason for such failures. Similarly, there seems to be a clear pattern where task-aware methods perform significantly worse on structured datasets than on natural ones. We hypothesize that this is due to the lack of adequate expert models for these domains. However, an in-depth analysis of these specific cases might be beneficial and insightful.

\textbf{Social impact.} We examine and propose a mixed model search strategy focusing solely on maximizing downstream accuracy. Applying such a search strategy blindly might enable hurtful biases present in pre-trained models, or in the datasets used to train the models upstream, to propagate into the models trained downstream. Quantifying the impact of transfer learning and different search strategies beyond accuracy is left as future work, whereby we encourage users of online pre-trained models to check for biases and fairness issues after fine-tuning on their custom datasets.

{
\small
\textbf{Acknowledgements.} We thank Jeremiah for the inspiration to work on a search functionality of TF-Hub models, and Josip for reviewing our paper internally. CZ and the DS3Lab gratefully acknowledge the support from the Swiss State Secretariat for Education, Research and Innovation (SERI)’s Backup Funding Scheme for European Research Council (ERC) Starting Grant TRIDENT (101042665), the Swiss National Science Foundation (Project Number 200021\_184628, and 197485), Innosuisse/SNF BRIDGE Discovery (Project Number 40B2-0\_187132), European Union Horizon 2020 Research and Innovation Programme (DAPHNE, 957407), Botnar Research Centre for Child Health, Swiss Data Science Center, Alibaba, Cisco, eBay, Google Focused Research Awards, Kuaishou Inc., Oracle Labs, Zurich Insurance, and the Department of Computer Science at ETH Zurich.
}

{\small
\bibliographystyle{ieee_fullname}
\bibliography{references}
}

\clearpage
\newpage

\appendix

\onecolumn

\section{Pre-Trained model details}
\label{app:ModelDetails}

\FloatBarrier

We provide a detailed list of all the used pre-trained models together with the dimension of their representations, the number of parameters, and the achieved ImageNet test accuracy (for those that the accuracy is known), in the following tables.

\begin{table}[!htb]
\centering
\begin{tabular}{@{}llll@{}}
\toprule
Model Name & Dim  & \# Params & ImageNet Accuracy \\ \midrule
inception\_v1/feature\_vector/4           & 1024 & 5'592'624   & 0.698             \\
inception\_v2/feature\_vector/4           & 1024 & 10'153'336  & 0.739             \\
inception\_v3/feature\_vector/4           & 2048 & 21'768'352  & 0.78              \\
inception\_resnet\_v2/feature\_vector/4   & 1536 & 54'276'192  & 0.804             \\
resnet\_v1\_50/feature\_vector/4          & 2048 & 23'508'032  & 0.752             \\
resnet\_v1\_101/feature\_vector/4         & 2048 & 42'500'160  & 0.764             \\
resnet\_v1\_152/feature\_vector/4         & 2048 & 58'143'808  & 0.768             \\
resnet\_v2\_50/feature\_vector/4          & 2048 & 23'519'360  & 0.756             \\
resnet\_v2\_101/feature\_vector/4         & 2048 & 42'528'896  & 0.77              \\
resnet\_v2\_152/feature\_vector/4         & 2048 & 58'187'904  & 0.778             \\
mobilenet\_v1\_100\_224/feature\_vector/4 & 1024 & 3'206'976   & 0.709             \\
mobilenet\_v2\_100\_224/feature\_vector/4 & 1280 & 2'223'872   & 0.718             \\
nasnet\_mobile/feature\_vector/4          & 1056 & 4'232'978   & 0.74              \\
nasnet\_large/feature\_vector/4           & 4032 & 84'720'150  & 0.827             \\
pnasnet\_large/feature\_vector/4          & 4320 & 81'736'668  & 0.829             \\ \bottomrule
\end{tabular}
\caption{ImageNet Classification Models -- All models are accessible by using the same prefix ``\url{https://tfhub.dev/google/imagenet/}'' in front of the model name.}
\end{table}

\begin{table}[!htb]
\centering
\begin{tabular}{@{}lll@{}}
\toprule
Model (Subset) & Dim  & \# Params \\ \midrule
Mode of transport  & 2048 & 23'807'702  \\
Geographical feature & 2048 & 23'807'702  \\
Structure & 2048 & 23'807'702  \\
Mammal & 2048 & 23'807'702  \\
Plant & 2048 & 23'807'702  \\
Material & 2048 & 23'807'702  \\
Home \& garden & 2048 & 23'807'702  \\
Flowering plant & 2048 & 23'807'702  \\
Sports equipment & 2048 & 23'807'702  \\
Dish & 2048 & 23'807'702  \\
Textile & 2048 & 23'807'702  \\
Shoe & 2048 & 23'807'702  \\
Bag & 2048 & 23'807'702  \\
Paper & 2048 & 23'807'702  \\
Snow & 2048 & 23'807'702  \\
\textit{Full JFT} & 2048 & 23'807'702  \\ \bottomrule
\end{tabular}
\caption{Expert Models -- The model name indicates the subset of JFT on which each model was trained~\cite{puigcerver2020experts}.}
\end{table}

\newpage
\begin{table}[!htb]
\centering
\begin{tabular}{@{}lll@{}}
\toprule
Model Name & Dim  & \# Params \\ \midrule
sup-100/1                 & 2048 & 23'500'352  \\
rotation/1                & 2048 & 23'500'352  \\
exemplar/1                & 2048 & 23'500'352  \\
relative-patch-location/1 & 2048 & 23'500'352  \\
jigsaw/1                  & 2048 & 23'500'352  \\
semi-rotation-10/1        & 2048 & 23'500'352  \\
sup-rotation-100/1        & 2048 & 23'500'352  \\
semi-exemplar-10/1        & 2048 & 23'500'352  \\
sup-exemplar-100/1        & 2048 & 23'500'352  \\
cond-biggan/1             & 1536 & 86'444'833  \\
uncond-biggan/1           & 1536 & 86'444'833  \\
wae-mmd/1                 & 128  & 23'779'136  \\
wae-gan/1                 & 128  & 23'779'136  \\
wae-ukl/1                 & 128  & 23'779'136  \\
vae/1                     & 128  & 23'779'136  \\ \bottomrule
\end{tabular}
\caption{VTAB Benchmark Models -- All models are accessible by using the model name and the prefix ``\url{https://tfhub.dev/vtab/}''.}
\end{table}

\clearpage
\newpage
\section{Limitation of correlation as evaluation score}
\label{app:limitationCorrelation}

Previous works suggests to perform a correlation analysis for choosing which model to transfer based on a ranking across the models~\cite{kornblith2019better, meiseles2020source}. We claim that this is not a suitable score in our setting of heterogenous pools, and in this section we explain the arguments in more details. We provide a simple example in which a correlation analysis fails compared to our notion of regret, which we see as an intuitive notion of failure in this setting.

We start by outlining two obvious dependencies between the two variants:
\begin{itemize}
    \item Having a perfect correlation (equal to $1$) results in zero regret,
    \item Having zero regret does not necessarily imply a perfect correlation.
\end{itemize}

The first statement follows by definition, whereas the second statement is justified by the following example: suppose that all models perform identically in terms of the fine-tune accuracy. In this case, every attribute (e.g.\ proxy task value, or ImageNet accuracy) would yield no correlation with respect to the fine-tune accuracy, although there is clearly zero regret for every imaginable strategy.

\begin{figure}[!htb]
\centering\captionsetup{width=1.0\linewidth}
\includegraphics[width=1.0\textwidth]{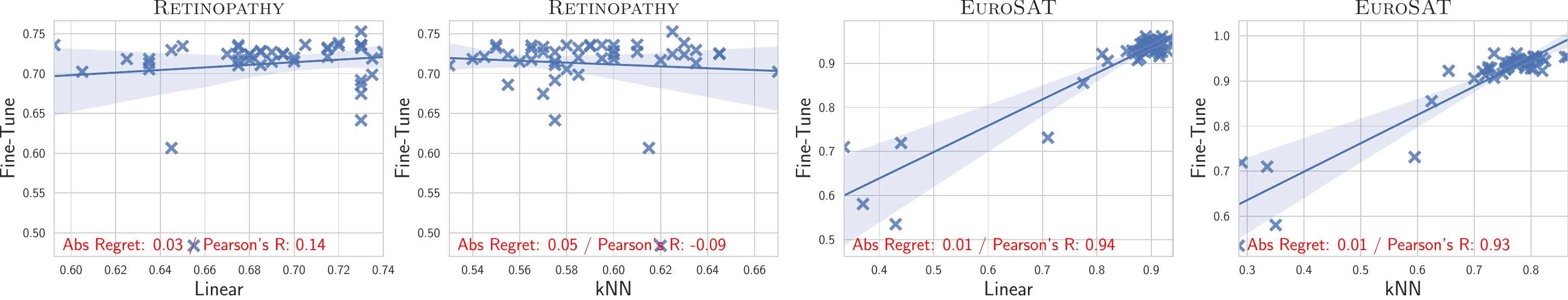}
\caption{Example correlation values for pool \all.}
\label{fig:correlation_examples_all}
\end{figure}

\begin{figure}[!htb]
\centering\captionsetup{width=1.0\linewidth}
\includegraphics[width=1.0\textwidth]{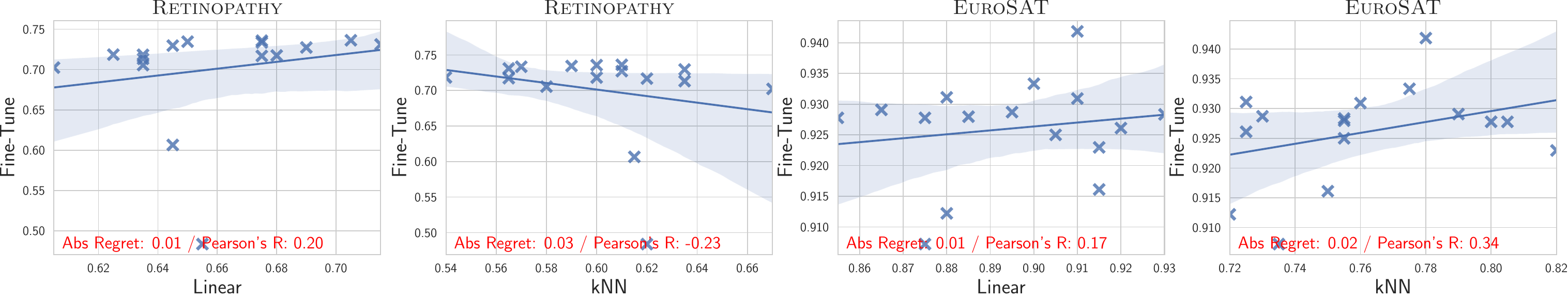}
\caption{Example correlation values for pool \jftslices.}
\label{fig:correlation_examples_expert}
\end{figure}

\paragraph{Implications.}
If we have a large pool of models with some outlier models that clearly outperform the others, which are of similar fine-tune accuracy, a search strategy should return one of those \textit{better} model, otherwise it will suffer from a large regret. On the other hand, this setting would usually have no rank nor linear correlation following the reasoning from before. If we restrict the same pool to models performing similarly, it will remain uncorrelated, but every search strategy will result in zero regret.  The same scenario holds if single outliers are performing worse than all other models.
Both cases are seen often in practice, especially for model pools containing experts. We highlight some examples of this phenomena in Figures~\ref{fig:correlation_examples_all} and \ref{fig:correlation_examples_expert}, together with some that have positive correlation.

\FloatBarrier

\clearpage \newpage
\section{Log-odds for the evaluation score}\label{app:logVSrel}

Following \cite{kornblith2019better}, we analyze a notion that differs from our definition of relative regret and delta between two strategies (cf. Equation~\ref{eq:rel_delta} and Section~\ref{sec:regret}). The idea is to compare two search strategies $m_1$ and $m_2$ by calculating the difference between $\textrm{logit}(s(m_1))$ and $\textrm{logit}(s(m_2))$  (the expected maximal logit-transformed test accuracy achieved by any model in the sets returned for both search strategies).
The logit transform is defined as $\textrm{logit}(p) = \log(p/(1-p)) = \textrm{sigmoid}^{-1}(p)$, also known as the log-odds of $p$. This transformation leads to the next definition of \textit{log-odds delta}:

\begin{equation}
    \label{eq:logodd_regret}
    \Tilde{\Delta}(m_1, m_2) := \log \left( \frac{s(m_1) - s(m_2)}{1-\min (s(m_1), s(m_2))} \right).
\end{equation}

Substituting $s(m_1)$ by the $\textsc{oracle}$ value from Equation~\ref{eq:regret} in Section~\ref{sec:regret}, and $s(m_2)$ by $s(m)$ leads to a new definition of \textit{log-odds regret} $\Tilde{r}(m)$.

These definitions are also incorporating the dataset difficulty and yield results very similar to our definition of the relative delta and relative regret in Section~\ref{sec:regret}. We now provide the analogous plots of the ones given in the main body of the paper, with the log-odds regret and log-odds delta instead of the relative regret and relative delta. We highlight the fact that, beside the change of the scale on the y-axis, all the findings given in the main body of the paper hold.

\FloatBarrier

\begin{figure}[!htb]
\centering\captionsetup{width=1.0\linewidth}
\includegraphics[width=1.0\textwidth]{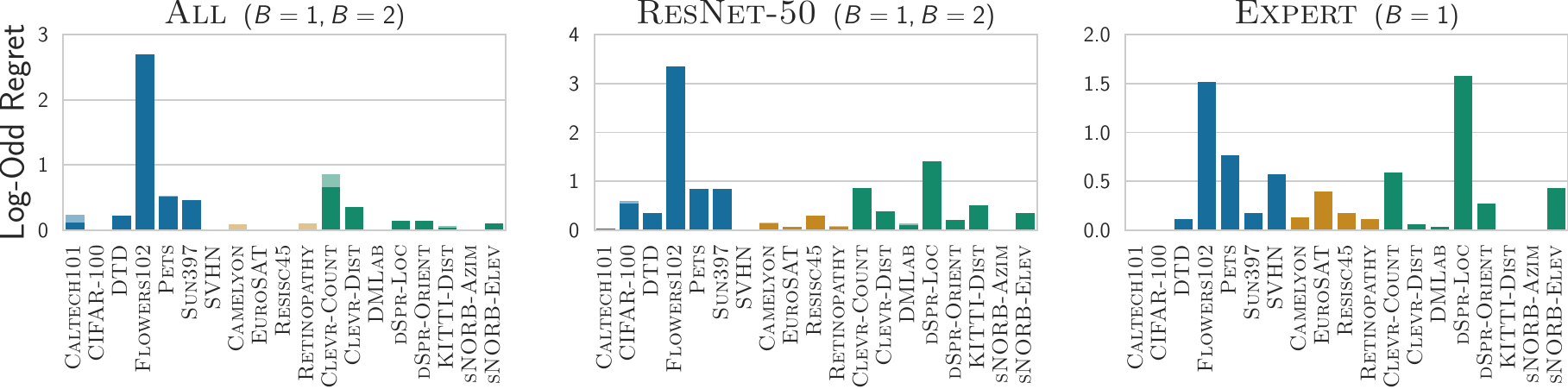}
\caption{Log-odds regret ($\Tilde{r}(m)$) with $B=1$ (transparent) and $B=2$ (solid) for the task-agnostic model search strategy.}
\label{fig:task_agnostic_fails_logodds}
\end{figure}

\begin{figure}[!htb]
\centering\captionsetup{width=1.0\linewidth}
\includegraphics[width=1.0\textwidth]{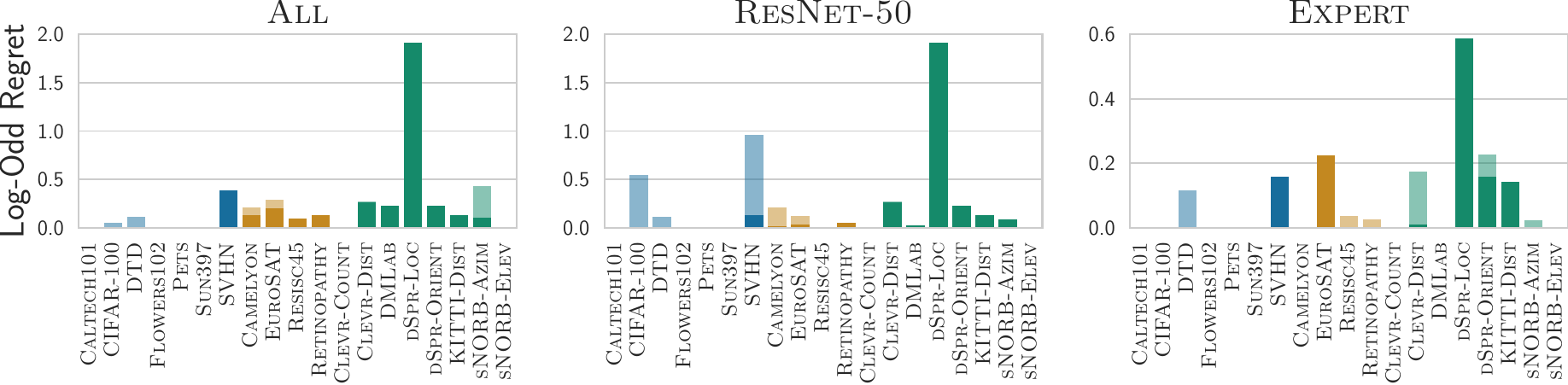}
\caption{Log-odds regret ($\Tilde{r}(m)$) for $B=1$ (transparent) and $B=2$ (solid) for the task-aware (linear) model search strategy.}
\label{fig:linear_fails_logodds}
\end{figure}

\clearpage
\newpage
\begin{figure}[!htb]
\centering\captionsetup{width=1.0\linewidth}
\includegraphics[width=1.0\textwidth]{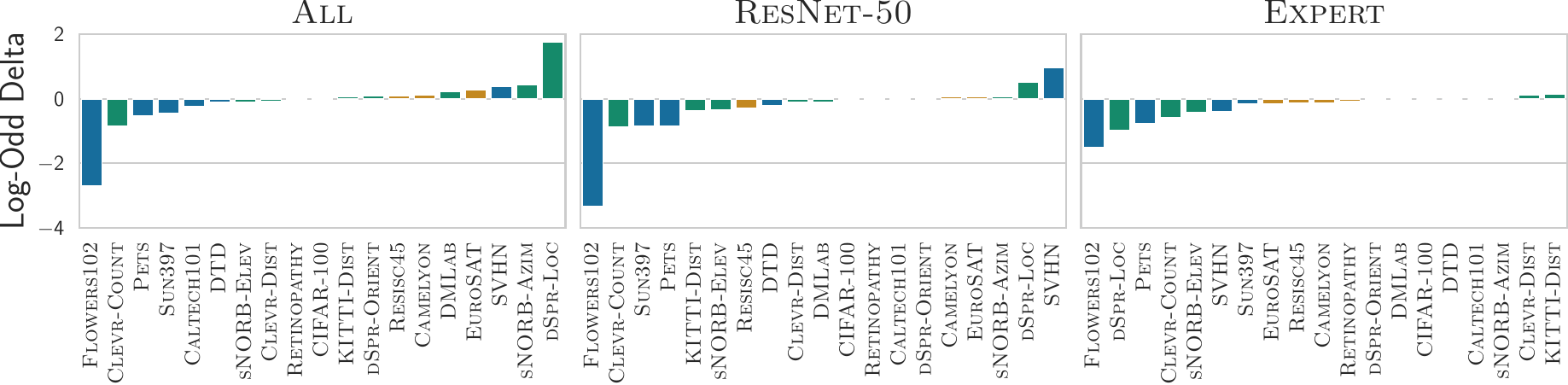}
\caption{Log-odds delta ($\Tilde{\Delta}(m_1, m_2)$) between task-agnostic (positive if better) and task-aware (linear) (negative if better) for $B=1$.}
\label{fig:task_aware_vs_task_agnostic_logodds}
\end{figure}

\begin{figure}[!htb]
\centering\captionsetup{width=1.0\linewidth}
\includegraphics[width=1.0\textwidth]{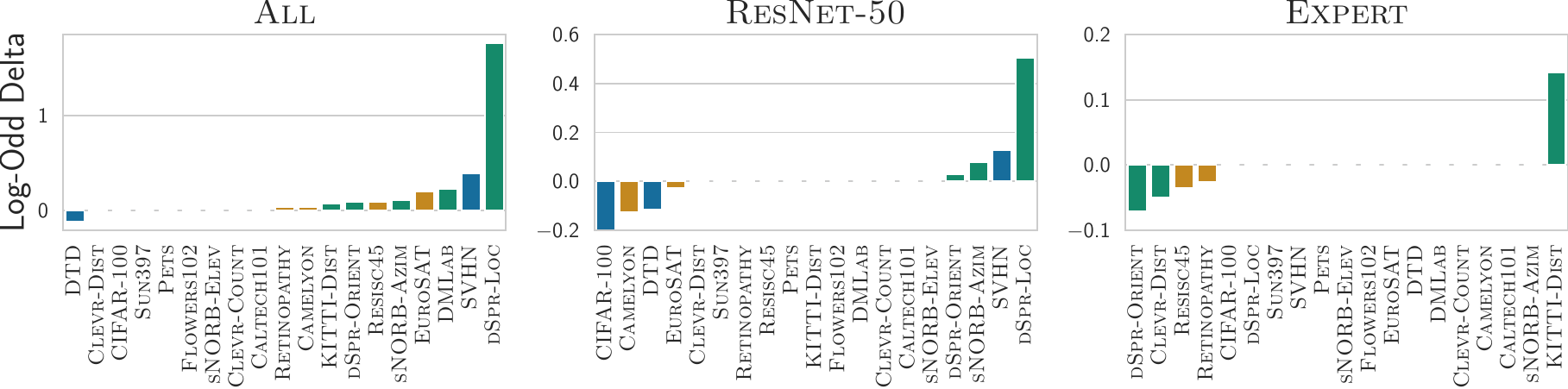}
\caption{Log-odds delta ($\Tilde{\Delta}(m_1, m_2)$) between hybrid linear (positive if better) and linear evaluation (negative if better) for $B=2$.}
\label{fig:hybrid_vs_linear_logodds}
\end{figure}

\FloatBarrier

\clearpage
\newpage

\section{Analysis for other pools.}\label{sec:app_other_pools}

In this sections we provide the plots and an analysis for the \restrdim\, and \imnetacc\, pools, both omitted from the main body of the paper due to space limitations.

We start by emphasizing that the results between the \restrdim\, and the \maxresnet\, pool, used in the main body of the paper, do not vary significantly. Most notably, the hybrid linear strategy is on par with the task-aware method, whereas the task-agnostic method suffers from high regret due to the lack of ability to pick expert models.

\begin{figure}[!htb]
\centering\captionsetup{width=1.0\linewidth}
\includegraphics[width=0.67\textwidth]{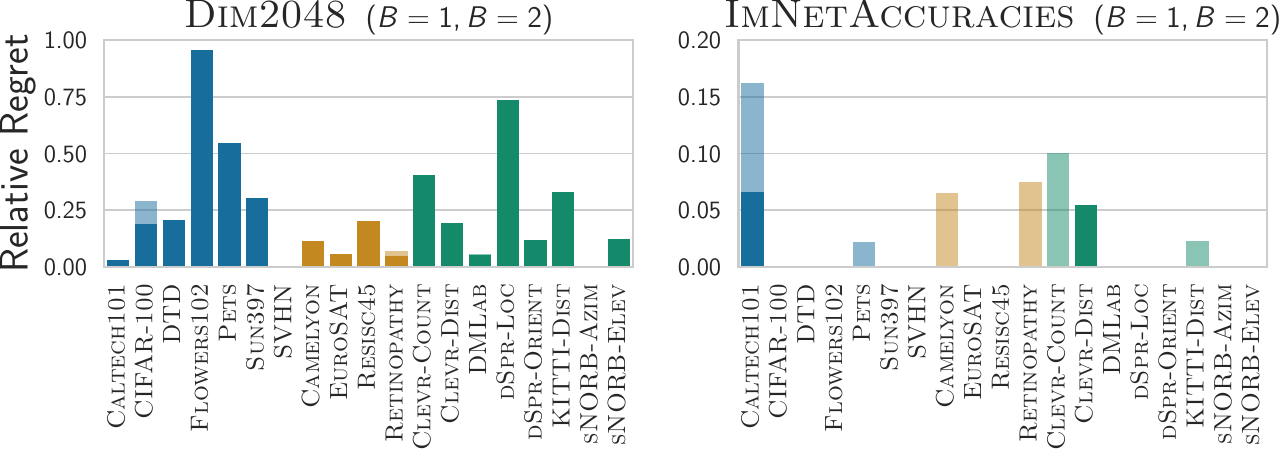}
\caption{Relative regret for the task-agnostic search strategy with $B=1$ (transparent) and $B=2$ (solid) on the pools \restrdim\, and \imnetacc.}
\label{fig:task_agnostic_fails_other_pools}
\vspace{-3mm}
\end{figure}

\begin{figure}[!htb]
\centering\captionsetup{width=1.0\linewidth}
\includegraphics[width=0.67\textwidth]{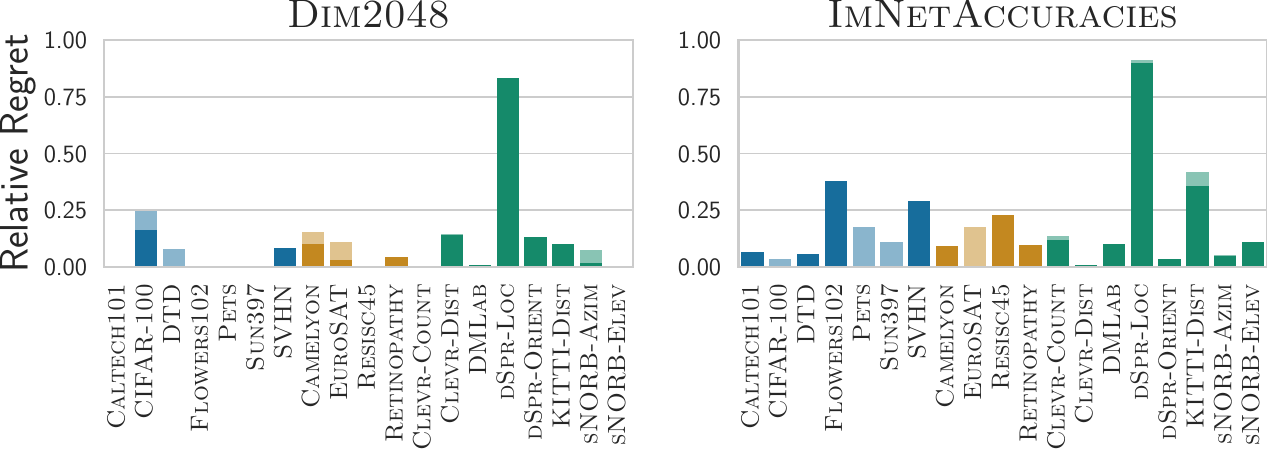}
\caption{Relative regret for the task-aware (linear) search strategy with $B=1$ (transparent) and $B=2$ (solid) on the pools \restrdim\, and \imnetacc.}
\label{fig:linear_fails_other_pools_relative}
\vspace{-3mm}
\end{figure}

\begin{figure}[!htb]
\centering\captionsetup{width=1.0\linewidth}
\includegraphics[width=0.67\textwidth]{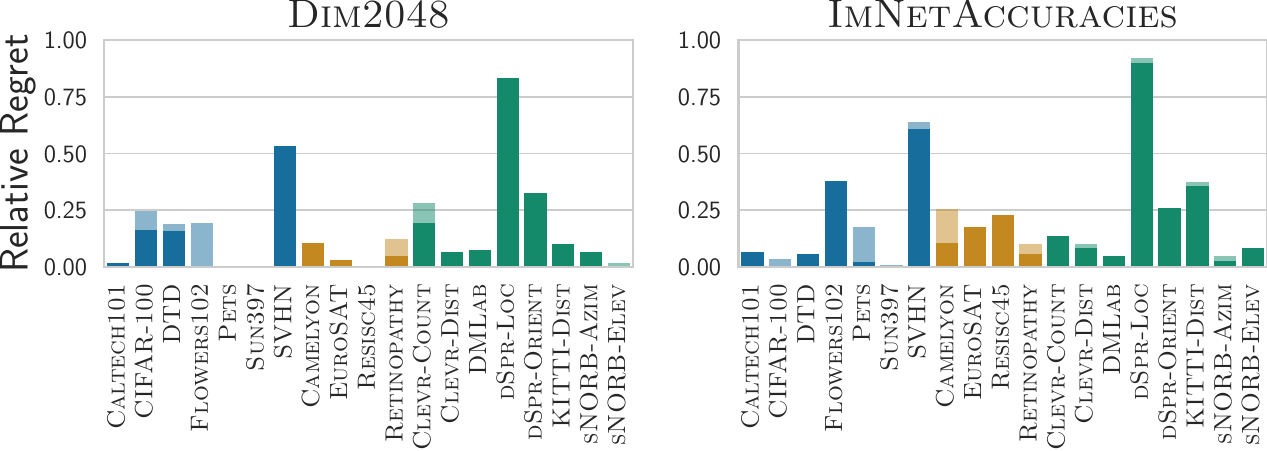}
\caption{Relative regret for the task-aware ($k$NN) search strategy with $B=1$ (transparent) and $B=2$ (solid) on the pools \restrdim\, and \imnetacc.}
\label{fig:knn_fails_other_pools_relative}
\end{figure}

When examining the performance of all strategies on the \imnetacc\, pool, presented in Figures~\ref{fig:task_agnostic_fails_other_pools}~(right), \ref{fig:linear_fails_other_pools_relative}~(right) and~\ref{fig:knn_fails_other_pools_relative}~(right), we observe that the task-agnostic strategy is able to pick the optimal model for 12 out of 19 datasets for $B=1$, and as many as 17 out of 19 for $B=2$. This clearly confirms the claim made by \cite{kornblith2019better} that \textit{better ImageNet models transfer better}. More surprisingly, we observe that both task-aware strategies (linear and $k$NN) fail consistently and, hence, result in high regret when being restricted to the \imnetacc\, pool only (cf. Figures~\ref{fig:linear_fails_other_pools_relative} and \ref{fig:knn_fails_other_pools_relative}).

\FloatBarrier

\clearpage\newpage

\begin{figure}[!htb]
\centering\captionsetup{width=1.0\linewidth}
\includegraphics[width=0.67\textwidth]{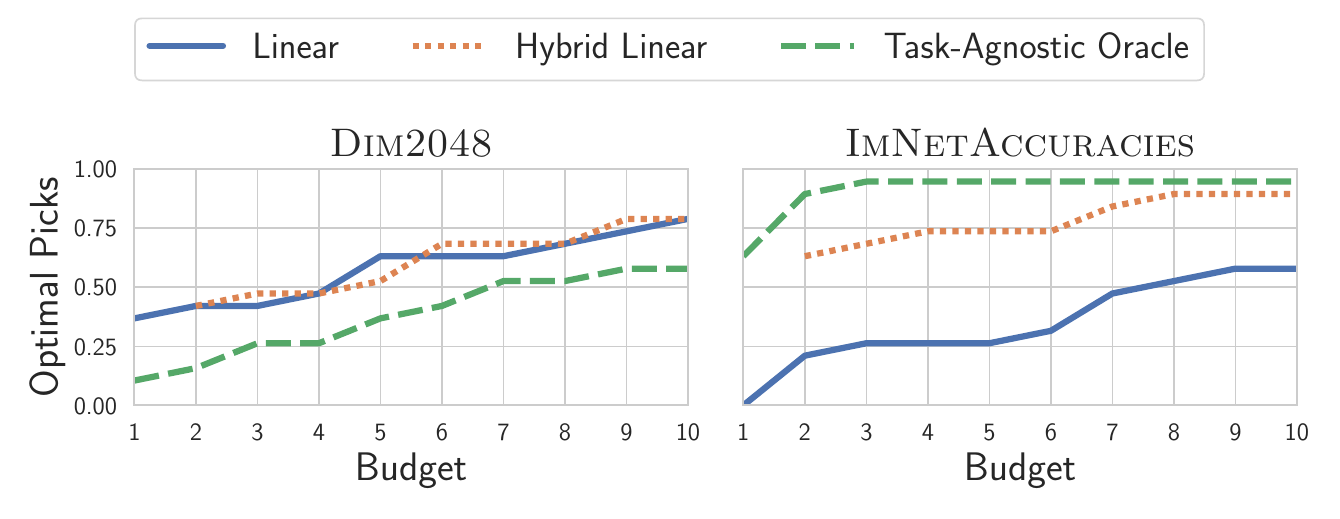}
\caption{Optimal picks for an increasing budget on \restrdim\, and \imnetacc.}
\label{fig:budget_vs_correct_picks_other_pools}
\end{figure}

\FloatBarrier

\clearpage
\newpage

\section{\texorpdfstring{$k$NN}{kNN} as a task-aware proxy}
\label{sec:app_full_task_aware}

\begin{figure}[!htb]
\centering\captionsetup{width=1.0\linewidth}
\includegraphics[width=1.0\textwidth]{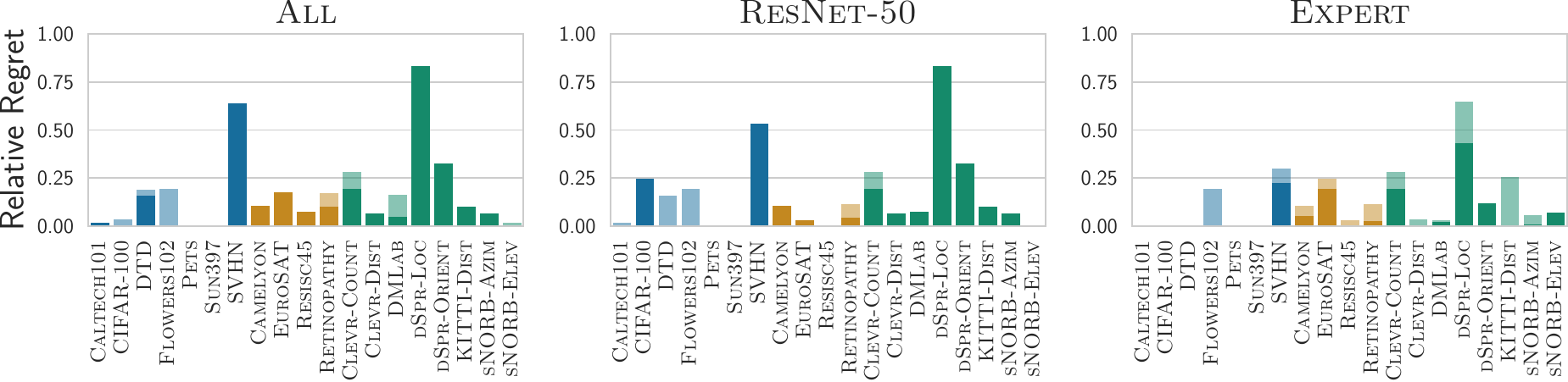}
\caption{Relative regret for the $k$NN search strategy with $B=1$ (transparent) and $B=2$ (solid).}
\label{fig:knn_fails_relative}
\end{figure}

\begin{figure}[!htb]
\centering\captionsetup{width=1.0\linewidth}
\includegraphics[width=0.95\textwidth]{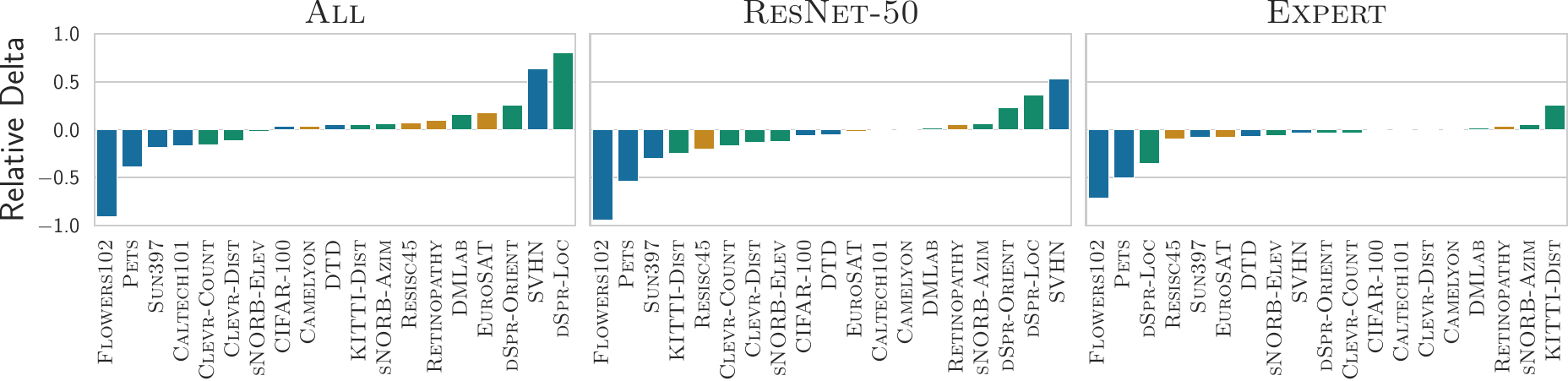}
\caption{Relative delta between the task-agnostic (positive if better) and the $k$NN task-aware search strategy (negative if better) for $B=1$.}
\label{fig:knn_vs_task_agnostic_relative}
\end{figure}

\begin{figure}[!htb]
\centering\captionsetup{width=1.0\linewidth}
\includegraphics[width=1.0\textwidth]{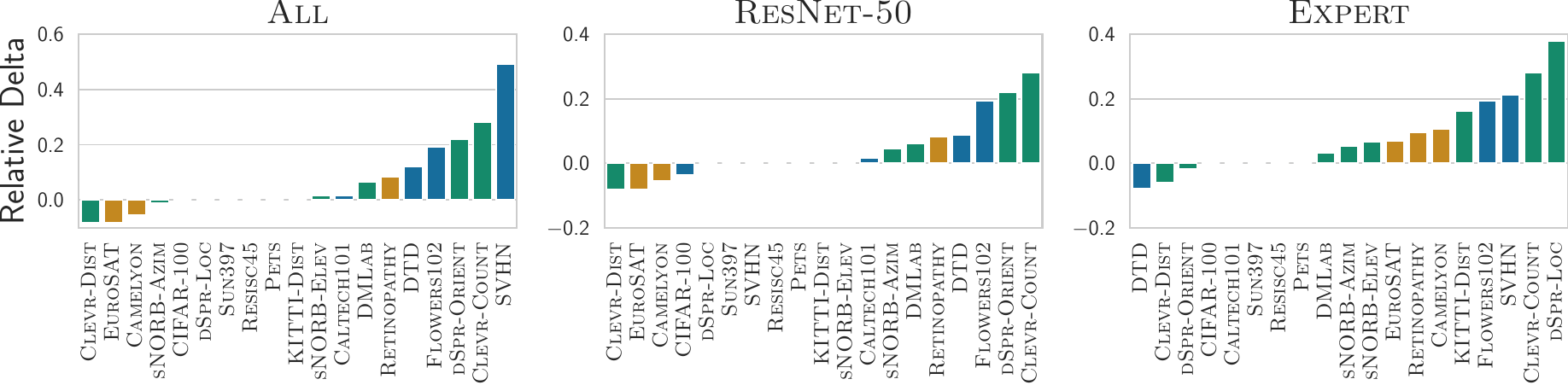}
\caption{Relative delta between the linear (positive if better) and the $k$NN  task-aware search strategy (negative if better) for $B=1$.}
\label{fig:linear_vs_knn_relative}
\end{figure}

\begin{figure}[!htb]
\centering\captionsetup{width=1.0\linewidth}
\includegraphics[width=0.95\textwidth]{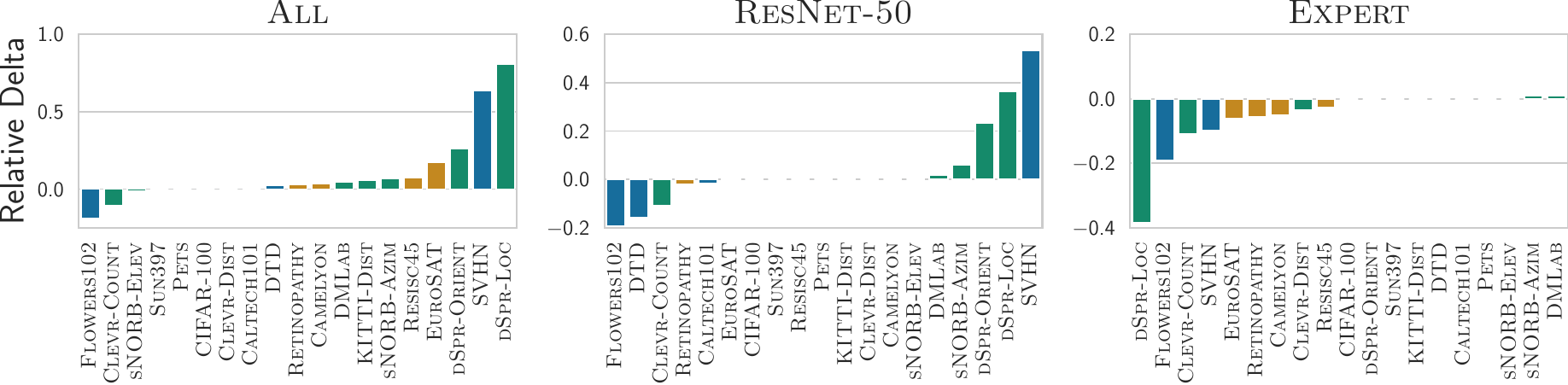}
\caption{Relative delta between the hybrid kNN (positive if better) and kNN search strategy (negative if better) for $B=2$.}
\label{fig:hybrid_vs_knn_relative}
\end{figure}

\begin{figure}[!htb]
\centering\captionsetup{width=1.0\linewidth}
\includegraphics[width=1.0\textwidth]{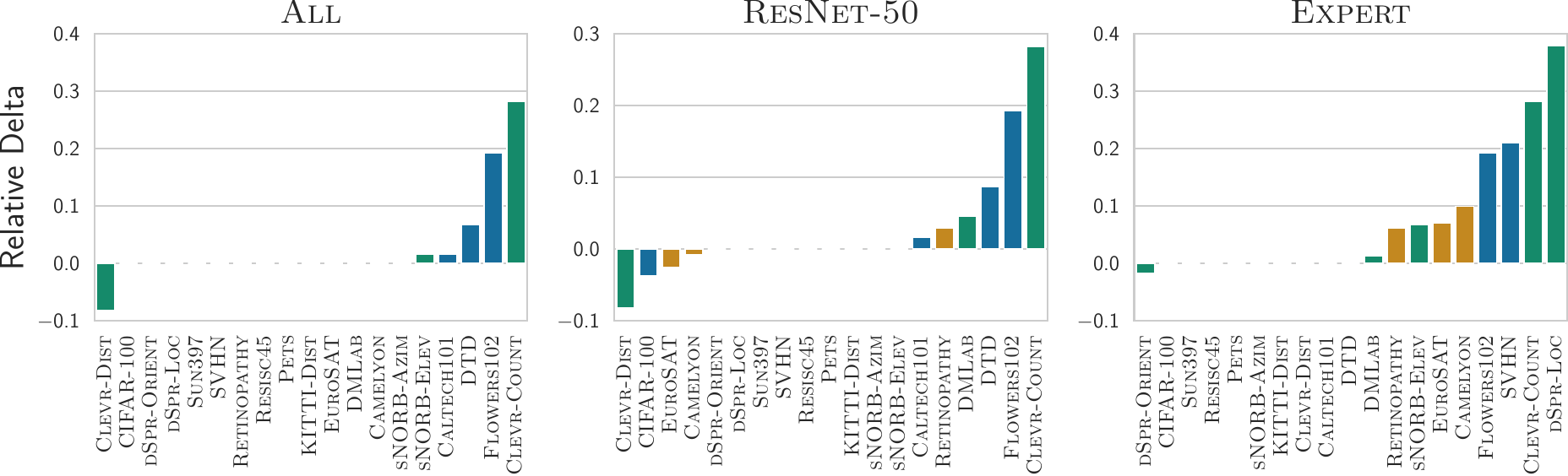}
\caption{Relative delta between the hybrid  Linear (positive if better) and hybrid kNN search strategy (negative if better) for $B=2$.}
\label{fig:hybrid_linear_vs_hybrid_knn_relative}
\end{figure}

In this section, we analyze the impact of choosing the $k$NN classifier accuracy as the choice for the proxy task, compared to the linear classifier accuracy described in the main body of the paper.
In practice, $k$NN might be favorable to a user since calculating the $k$NN classifier accuracy with respect to a relatively small test set can be orders of magnitude faster compared to training a linear classifier, which might be sensitive to the choice of optimal hyper-parameters. A theoretical and empirical analysis of the computing performance of these two proxies is out of the scope of this work, as we are mainly interested in the comparison in terms of the model-search capability of either of these.
In general, the major claims on the performance of linear as a task-aware strategy also apply to $k$NN. The latter also mainly fails on structured datasets across all pools, as visible in Figure~\ref{fig:knn_fails_relative}. Similarly to the linear task, $k$NN is on par with the task-agnostic strategy on the \all\, pool, but clearly outperforms it on the \maxresnet\, and \jftslices\, pools, as visible in Figure~\ref{fig:knn_vs_task_agnostic_relative}. By further comparing $k$NN to the linear proxy task, we realize that $k$NN performs worse than linear on half of the datasets across the three different dataset groups, whilst being on par with it on the other half (cf. Figure~\ref{fig:linear_vs_knn_relative}). Finally, by choosing $k$NN as the task-aware part for the hybrid strategy and comparing it to the task-aware ($k$NN) strategy with a budget of $B=2$ in Figure~\ref{fig:hybrid_vs_knn_relative}, we see an increase of performance on the \all\, pool, no clear winner on the restricted \maxresnet\, pool, but higher regret on the \jftslices\, pool. Unsurprisingly, this version of hybrid strategy also performs slightly worse compared than the one with a linear proxy across all the pools (cf. Figure~\ref{fig:hybrid_linear_vs_hybrid_knn_relative}).

\FloatBarrier

\newpage

\clearpage
\FloatBarrier

\clearpage\newpage

\section{On the impact of the dimension on \texorpdfstring{$k$NN}{kNN}}\label{app:knn_vs_dim}

\begin{figure}[!htb]
\centering\captionsetup{width=1.0\linewidth}
\includegraphics[width=1.0\textwidth]{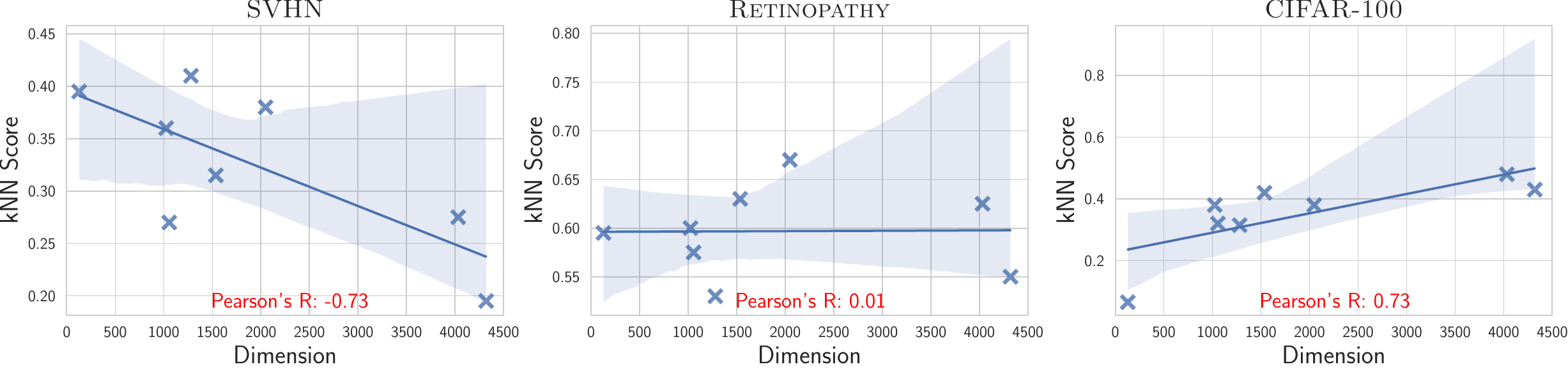}
\caption{Three examples of datasets in which the analysis of the dimension of the representation compared to the resulting $k$NN scores results in a negative correlation \textbf{(left)}, no correlation at all \textbf{(middle)}, and a positive correlation \textbf{(right)}.}
\label{fig:knn_vs_dim_coorelation}
\end{figure}

As described in Section~\ref{sec:exp_ablation}, in this section we show that there is no signification correlation (positive or negative) between the $k$NN classifier accuracy and the dimension of the representation that it is evaluated on. We see this by running a linear correlation analysis between the dimension of the each representation and the achieved $k$NN classifier accuracy. In order to have a single point for each possible dimension, and to avoid an over-representation of the expert models, which have all the same dimension, for each dimension we selected the model that achieves the highest $k$NN accuracy. We do this for all pairs of dimensions and datasets. In Figure~\ref{fig:knn_vs_dim_coorelation} we present three hand-picked datasets that achieve (a) the highest anti-correlation value, (b) the lowest absolute correlation value, and (c) the highest correlation value.

\FloatBarrier

\FloatBarrier

\clearpage
\newpage

\section{Budget per method}
\label{sec:appendix:budget_per_method}

In this section, we report the budget each method requires in order to achieve zero regret per pool. Notice that the strategy ``Oracle'' refers to the task-agnostic oracle which ranks models based on their achieved average accuracy over all datasets. Even though this is not practical, it enables us to have a task-agnostic method that is able to achieve zero regret eventually as every model (even an expert) is included in this ranking. We split the results by the dataset types and notice that there are some clear patterns between a pool, a dataset type and the required budget for task-aware or task-agnostic methods. For instance, one observes that the linear strategy performs well on all the natural datasets across all the pools except \imnetacc. Structured and specialized datasets seem to be harder for this proxy task, except for the \jftslices\, pool. Finally, $k$NN consistently performs slightly worse than the linear proxy across all pools.

\FloatBarrier

\begin{table}[!htb]
\centering
\begin{tabular}{@{}rccc|ccc|ccc@{}}
\toprule
            & \multicolumn{3}{c|}{\all} & \multicolumn{3}{c|}{\maxresnet} & \multicolumn{3}{c}{\jftslices} \\ 
Dataset     & Oracle  & Linear  & kNN & Oracle    & Linear    & kNN   & Oracle   & Linear   & kNN  \\ \midrule
$\textsc{Caltech101}$  \naturalsym & 5        & 1          & 4       & 3              & 1                & 2             & 1           & 1             & 1          \\
$\textsc{CIFAR-100}$ \naturalsym  & 1        & 2          & 2       & 1              & 2                & 6             & 1           & 1             & 1          \\
$\textsc{DTD}$   \naturalsym      & 9        & 2          & 3       & 6              & 2                & 2             & 3           & 2             & 1          \\
$\textsc{Flowers102}$ \naturalsym & 26       & 1          & 2       & 20             & 1                & 2             & 11          & 1             & 2          \\
$\textsc{Pets}$  \naturalsym      & 13       & 1          & 1       & 9              & 1                & 1             & 5           & 1             & 1          \\
$\textsc{Sun397}$   \naturalsym   & 7        & 1          & 1       & 5              & 1                & 1             & 2           & 1             & 1          \\
$\textsc{SVHN}$   \naturalsym     & 1        & 23         & 37      & 2              & 9                & 13            & 15          & 13            & 15         \\ \midrule
$\textsc{Camelyon}$ \specializedsym   & 23       & 5          & 9       & 17             & 4                & 8             & 9           & 1             & 4          \\
$\textsc{EuroSAT}$  \specializedsym   & 1        & 4          & 19      & 14             & 16               & 7             & 4           & 5             & 5          \\
$\textsc{Resisc45}$  \specializedsym  & 1        & 35         & 34      & 3              & 1                & 1             & 2           & 2             & 2          \\
$\textsc{Retinopathy}$ \specializedsym& 2        & 4          & 8       & 23             & 14               & 9             & 12          & 2             & 7          \\ \midrule
$\textsc{Clevr-Count}$ \structuredsym & 7        & 1          & 10      & 5              & 1                & 9             & 2           & 1             & 6          \\
$\textsc{Clevr-Dist}$ \structuredsym & 44       & 5          & 4       & 35             & 5                & 4             & 10          & 6             & 2          \\
$\textsc{DMLab}$  \structuredsym     & 1        & 18         & 29      & 3              & 8                & 4             & 7           & 1             & 6          \\
$\textsc{dSpr-Loc}$  \structuredsym  & 9        & 25         & 21      & 6              & 19               & 17            & 3           & 3             & 4          \\
$\textsc{dSpr-Orient}$ \structuredsym & 30       & 9          & 21      & 23             & 8                & 19            & 12          & 5             & 12         \\
$\textsc{KITTI-Dist}$ \structuredsym & 3        & 8          & 14      & 1              & 7                & 10            & 1           & 3             & 2          \\
$\textsc{sNORB-Azim}$ \structuredsym & 1        & 45         & 25      & 32             & 3                & 5             & 1           & 2             & 15         \\
$\textsc{sNORB-Elev}$ \structuredsym & 37       & 1          & 2       & 30             & 1                & 1             & 12          & 1             & 5          \\ \bottomrule
\end{tabular}
\caption{Budget required to achieve zero regret per datataset and strategy on the pools \all, \maxresnet\, and \jftslices.}
\end{table}

\clearpage
\newpage
\begin{table}[!htb]
\centering
\begin{tabular}{@{}rccc|ccc@{}}
\toprule
            & \multicolumn{3}{c|}{\restrdim} & \multicolumn{3}{c}{\imnetacc} \\ 
Dataset     & Oracle  & Linear  & kNN & Oracle    & Linear   \\ \midrule
$\textsc{Caltech101}$ \naturalsym & 3            & 1              & 4           & 3                    & 7                      & 12                  \\
$\textsc{CIFAR-100}$  \naturalsym & 1            & 5              & 10          & 1                    & 2                      & 2                   \\
$\textsc{DTD}$   \naturalsym      & 7            & 2              & 3           & 2                    & 7                      & 9                   \\
$\textsc{Flowers102}$ \naturalsym & 24           & 1              & 2           & 1                    & 14                     & 15                  \\
$\textsc{Pets}$  \naturalsym      & 11           & 1              & 1           & 2                    & 2                      & 3                   \\
$\textsc{Sun397}$  \naturalsym    & 5            & 1              & 1           & 1                    & 2                      & 2                   \\
$\textsc{SVHN}$  \naturalsym      & 2            & 12             & 15          & 1                    & 14                     & 15                  \\ \midrule
$\textsc{Camelyon}$  \specializedsym  & 21           & 5              & 9           & 2                    & 7                      & 8                   \\
$\textsc{EuroSAT}$  \specializedsym   & 17           & 21             & 8           & 1                    & 2                      & 8                   \\
$\textsc{Resisc45}$  \specializedsym  & 3            & 1              & 1           & 1                    & 14                     & 14                  \\
$\textsc{Retinopathy}$ \specializedsym & 6            & 12             & 6           & 2                    & 3                      & 4                   \\ \midrule
$\textsc{Clevr-Count}$ \structuredsym& 5            & 1              & 10          & 2                    & 12                     & 12                  \\
$\textsc{Clevr-Dist}$ \structuredsym & 42           & 5              & 4           & 14                   & 6                      & 4                   \\
$\textsc{DMLab}$  \structuredsym     & 9            & 10             & 8           & 1                    & 9                      & 12                  \\
$\textsc{dSpr-Loc}$  \structuredsym  & 7            & 25             & 21          & 1                    & 15                     & 13                  \\
$\textsc{dSpr-Orient}$ \structuredsym & 28           & 9              & 21          & 1                    & 8                      & 12                  \\
$\textsc{KITTI-Dist}$ \structuredsym & 1            & 8              & 13          & 2                    & 13                     & 15                  \\
$\textsc{sNORB-Azim}$ \structuredsym & 37           & 4              & 7           & 1                    & 14                     & 9                   \\
$\textsc{sNORB-Elev}$ \structuredsym & 35           & 1              & 2           & 1                    & 15                     & 15                  \\ \bottomrule
\end{tabular}
\caption{Budget required to achieve zero regret per datataset and strategy on the pools \restrdim\, and \imnetacc.}
\end{table}

\FloatBarrier
\clearpage
\newpage

\section{Task2Vec results}
\label{sec:appendix:task2vec}

As described in Section~\ref{sec:background} and \ref{sec:other_related_work} of the main part of this work, we report some preliminary results on extending our model search formulation to meta-learned task-aware search strategies. In this section we look at Task2Vec~\cite{achille2019task2vec}, the predominant method in this area. Following the experiments by the authors of Task2Vec, we perform our evaluation in a leave-one-out (LOO) fashion. Concretely, for each dataset in the VTAB benchmark, we run the evaluation by taking the other 18 datasets as the benchmark set of datasets for the meta-learning part of the search strategy. We use the code provided with the Task2Vec paper\footnote{\url{https://github.com/awslabs/aws-cv-task2vec}} to calculate the Task2Vec representations on all VTAB datasets with the \texttt{ResNet34} probe network and the \texttt{variational} method to get the Fisher information matrix (FIM) for each task. In order to determine the nearest task for each dataset in the VTAB benchmark, we use the normalized cosine function, which translates to $d_{sym}(F_a, F_b)$ in the related work. We are not able to use the asymmetric distance function $d_{asym}(t_a \rightarrow t_b)$ proposed by the authors of Task2Vec due to the lack of generalist task in the benchmark set, which would yield the trivial embedding.

\begin{figure}[!htb]
\centering\captionsetup{width=1.0\linewidth}
\includegraphics[width=1.0\textwidth]{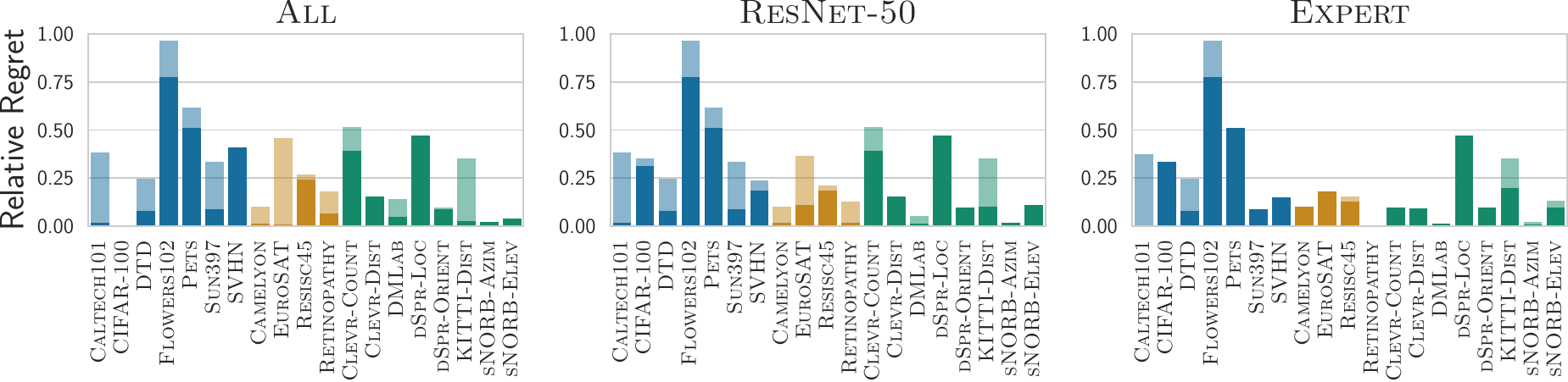}
\caption{Relative regret for the Task2Vec search strategy with $B=1$ (transparent) and $B=2$ (solid).}
\label{fig:task2vec_fails_relative}
\end{figure}

\begin{figure}[!htb]
\centering\captionsetup{width=1.0\linewidth}
\includegraphics[width=0.95\textwidth]{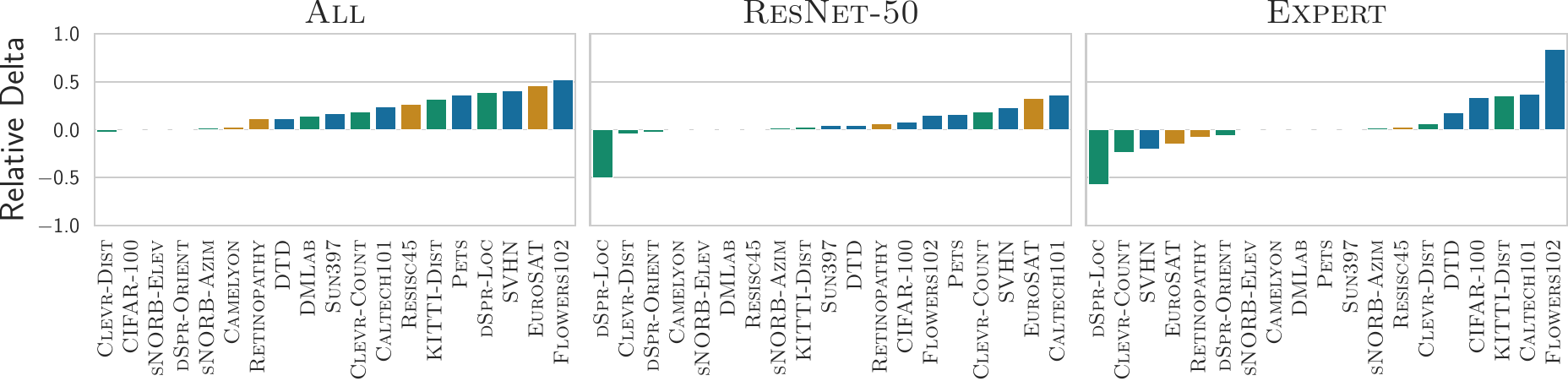}
\caption{Relative delta between the task-agnostic (positive if better) and the Task2Vec meta-learned task-aware search strategy (negative if better) for $B=1$.}
\label{fig:task2vec_vs_task_agnostic_relative}
\end{figure}

\begin{figure}[!htb]
\centering\captionsetup{width=1.0\linewidth}
\includegraphics[width=0.95\textwidth]{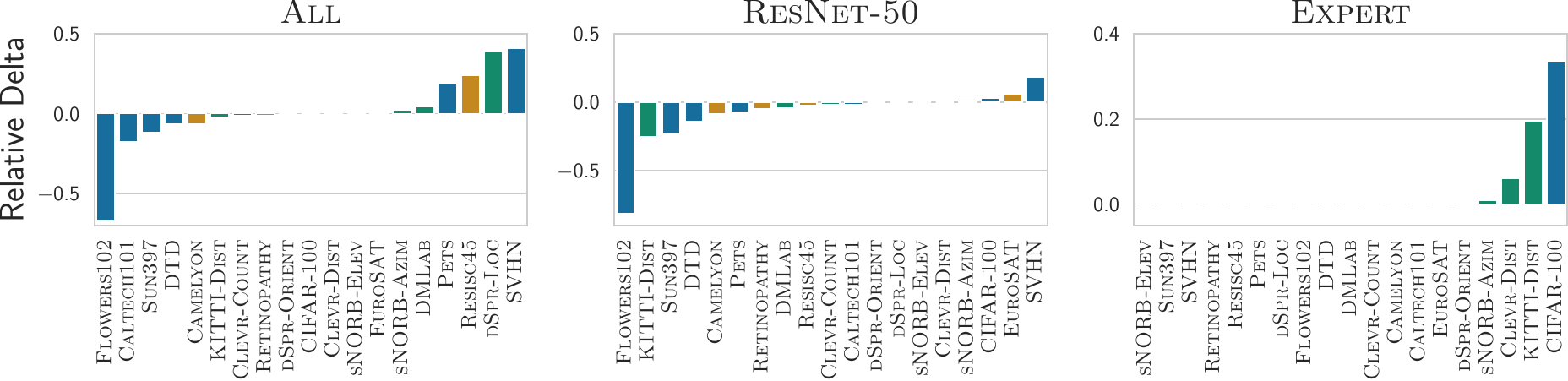}
\caption{Relative delta between the hybrid Task2Vec (positive if better) and Task2Vec search strategy (negative if better) for $B=2$.}
\label{fig:hybrid_vs_task2vec_relative}
\end{figure}

\paragraph{Results.}
We report the nearest task for each VTAB dataset in Table~\ref{tbl:nearest_vtab}. Notice that the nearest tasks for natural and structured datasets often belong to the same category.
When analyzing the relative regret from selecting the two best models from the nearest task to fine-tune across different pools in Figure~\ref{fig:task2vec_fails_relative}, we realize that, unsurprisingly, tasks such as \textsc{Flowers102}, where the nearest task seems to not be similar enough, suffer from a high regret of not finding the best expert model. Other tasks, such as \textsc{EuroSAT}, can benefit from picking the two best models instead of only one, a corner case beyond the initial analysis of the Task2Vec paper. Furthermore, we remark a high discrepancy between the regrets for tasks which are symmetrically the nearest tasks of each other (e.g.\ \textsc{dSpr-Loc} and \textsc{dSpr-Orient}).
When comparing the relative regret between the task-agnostic selection and the meta-learned task-aware strategy for a fixed budget of $B=1$, presented in Figure~\ref{fig:task2vec_vs_task_agnostic_relative}, we see that the simple task-agnostic strategy is on par, or better by a small margin, than the computational more demanding approach on most datasets.
This findings are supported by the authors of Task2Vec in Figure~3 in~\cite{achille2019task2vec}, where the expert selection procedure is outperformed by the simple generalist approach in 15 out of 50 tasks, sometimes by a large margin.
When increasing the budget to $B=2$ and comparing the hybrid Task2Vec strategy to the Task2Vec strategy, the benefits vanish and hybrid Task2Vec does not yield any significant improvements except for the \jftslices\, pool (cf. Figure~\ref{fig:hybrid_vs_task2vec_relative}).

\paragraph{Limitations and future work.}
Notice that we only inspect the best model(s) from the single nearest task. For $B>1$, one could also look at the second nearest task. Preliminary results did not show any improvements for $B=2$. Extending and designing new search strategies for meta-learned task-aware search and larger budgets is beyond the scope of this work. Furthermore, when exploring this area, one should carefully analyze the impact of having different sets of benchmark dataset. For instance, if a very similar dataset to the user's dataset is in the set of benchmark tasks, naturally the meta-learned task-aware will be very strong. Finally, notice that the asymmetric distance function proposed by the authors of Task2Vec, despite not being applicable to our setting, has strong resembles to the idea of balancing the search between picking a generalist (in their words, falling back to the trivial embedding), and the specialist model originating from the nearest task in the task embedding space. Enabling such a differentiation across all model search strategies without having to increase the budget offers an interesting research question for the future.

\FloatBarrier
\clearpage
\newpage

\begin{table}[!htb]
\centering
\begin{tabular}{@{}r|l@{}}
\toprule
Dataset     & Nearest Task \\ \midrule
$\textsc{Caltech101}$  \naturalsym & $\textsc{Sun397}$   \naturalsym \\
$\textsc{CIFAR-100}$ \naturalsym  & $\textsc{EuroSAT}$  \specializedsym \\
$\textsc{DTD}$   \naturalsym      & $\textsc{Sun397}$   \naturalsym \\
$\textsc{Flowers102}$ \naturalsym & $\textsc{Sun397}$   \naturalsym \\
$\textsc{Pets}$  \naturalsym      & $\textsc{Caltech101}$  \naturalsym \\
$\textsc{Sun397}$   \naturalsym   & $\textsc{Caltech101}$  \naturalsym \\
$\textsc{SVHN}$   \naturalsym     & $\textsc{sNORB-Elev}$ \structuredsym \\ \midrule
$\textsc{Camelyon}$ \specializedsym   & $\textsc{KITTI-Dist}$ \structuredsym \\
$\textsc{EuroSAT}$  \specializedsym   & $\textsc{Camelyon}$ \specializedsym \\
$\textsc{Resisc45}$  \specializedsym  & $\textsc{DTD}$   \naturalsym \\
$\textsc{Retinopathy}$ \specializedsym & $\textsc{sNORB-Elev}$ \structuredsym \\ \midrule
$\textsc{Clevr-Count}$ \structuredsym & $\textsc{Clevr-Dist}$ \structuredsym \\
$\textsc{Clevr-Dist}$ \structuredsym & $\textsc{Clevr-Count}$ \structuredsym \\
$\textsc{DMLab}$  \structuredsym     & $\textsc{KITTI-Dist}$ \structuredsym  \\
$\textsc{dSpr-Loc}$  \structuredsym  & $\textsc{dSpr-Orient}$ \structuredsym \\
$\textsc{dSpr-Orient}$ \structuredsym & $\textsc{dSpr-Loc}$  \structuredsym \\
$\textsc{KITTI-Dist}$ \structuredsym & $\textsc{Camelyon}$ \specializedsym \\
$\textsc{sNORB-Azim}$ \structuredsym & $\textsc{sNORB-Elev}$ \structuredsym \\
$\textsc{sNORB-Elev}$ \structuredsym & $\textsc{sNORB-Azim}$ \structuredsym \\ \bottomrule
\end{tabular}
\caption{Nearest tasks using Task2Vec for each VTAB dataset.}
\label{tbl:nearest_vtab}
\end{table}

\clearpage

\FloatBarrier
\clearpage
\newpage

\section{Early stopping as a proxy}
\label{sec:appendix:early_stopping}

As shown in this work, using a linear classifier on top of frozen representation as a proxy for picking which model to fine-tune can yield high regret. As a natural attempt to bypass this limitation, we define a computationally more demanding task-aware proxy method by fine-tuning each network, but only for a very small number of iterations, and use the resulting score as a proxy for performing the model search. We focus on a setting where we stay as close to the downstream training scenario described in Section~\ref{sec:background} and \ref{sec:experimentalsetup} as possible. Concretely, we initialize the new head as a zero-valued vector, run SGD for 10 iterations with a batch size of 512 (yielding 5 epochs), and use the largest validation accuracy achieved over the two choices of learning rates ($0.1$ and $0.01$). For robustness, we run this procedure 5 times and take the median validation accuracy. When inspecting the performance of ResNet50 pre-trained architectures, we realize that gradually increasing the learning rate over multiple iterations results in large variance across the different runs. Omitting such a ``warm-up'' strategy in the few iterations scenario yields more stable results.

\begin{figure}[!htb]
\centering\captionsetup{width=1.0\linewidth}
\includegraphics[width=1.0\textwidth]{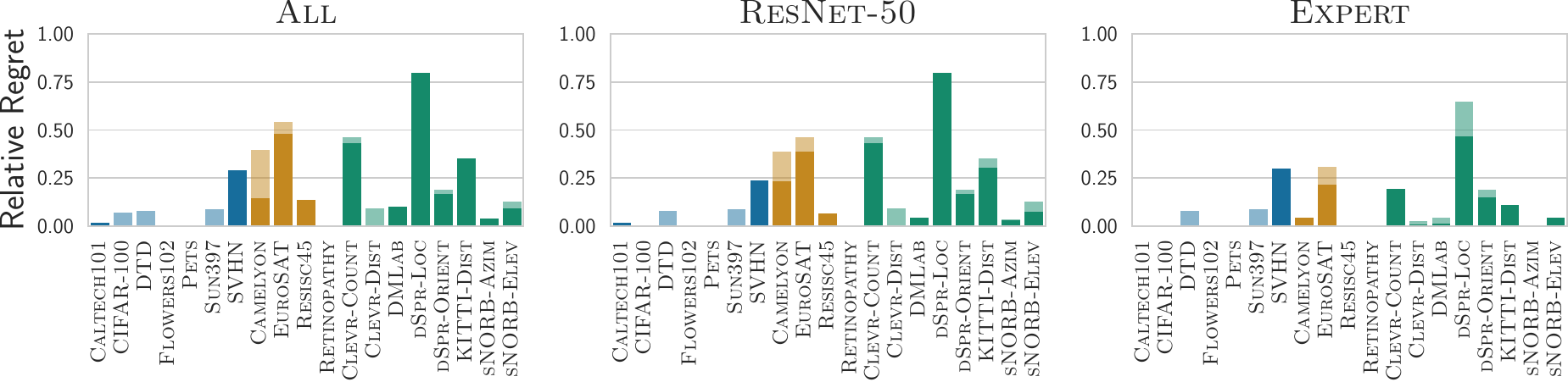}
\caption{Relative regret for the EarlyStop search strategy with $B=1$ (transparent) and $B=2$ (solid).}
\label{fig:earlystop_fails_relative}
\end{figure}

\begin{figure}[!htb]
\centering\captionsetup{width=1.0\linewidth}
\includegraphics[width=0.95\textwidth]{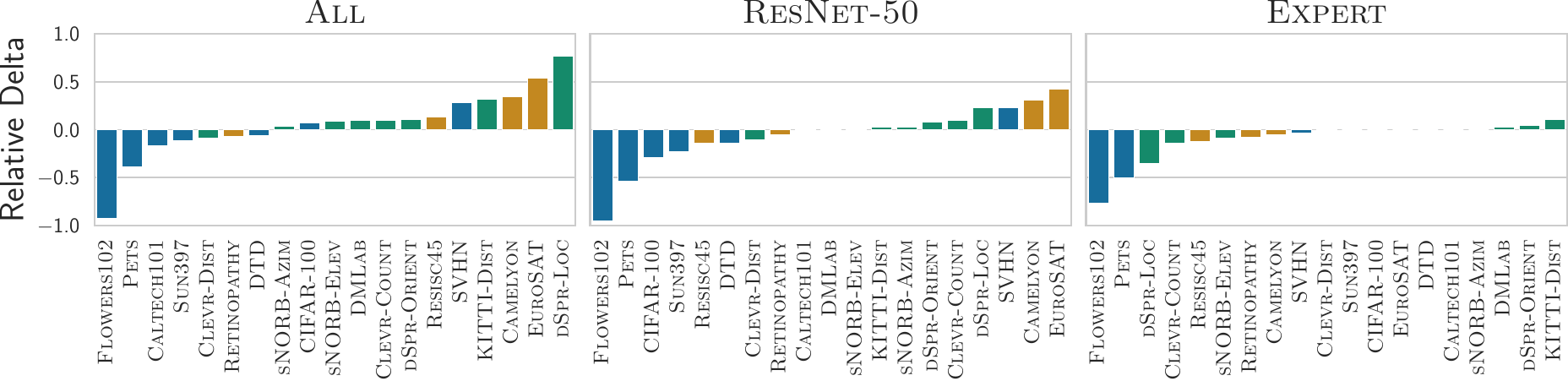}
\caption{Relative delta between the task-agnostic (positive if better) and the EarlyStop task-aware search strategy (negative if better) for $B=1$.}
\label{fig:earlystop_vs_task_agnostic_relative}
\end{figure}

\begin{figure}[!htb]
\centering\captionsetup{width=1.0\linewidth}
\includegraphics[width=0.95\textwidth]{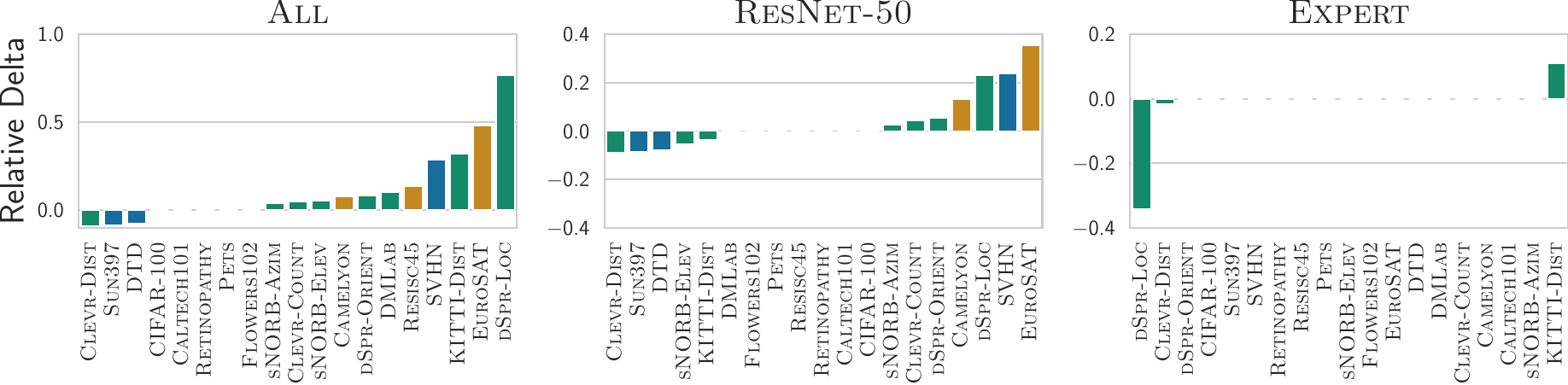}
\caption{Relative delta between the hybrid EarlyStop (positive if better) and EarlyStop search strategy (negative if better) for $B=2$.}
\label{fig:hybrid_vs_earlystop_relative}
\end{figure}

\paragraph{Results.}
The results in Figures~\ref{fig:earlystop_fails_relative}, \ref{fig:earlystop_vs_task_agnostic_relative} and \ref{fig:hybrid_vs_earlystop_relative}, follow the ones using the linear proxy on top of frozen representations. In fact, despite being computationally more demanding, fine-tuning for a few iterations does not bypass the limitations of using the simpler linear classifier proxy in our setting. We believe that the reasons are mainly due to the complex hyper-parameter search one should perform especially in the presence of varying architectures. When having only a few iterations, running any warm-up phase is non-trivial and thus selecting the \textit{best} initial learning rate becomes crucial. Preliminary evaluations showed that limiting the number of fine-tune iterations increases the impact and sensitivity to hyper-parameters. When comparing the validation accuracies between early stopped and fully fine-tuned models, some combination of pre-trained models and dataset require a small initial learning rate to not overshoot in the limited iterations scenario,  whereas others would require a larger learning rate in order to be able to increase the accuracy significantly. Further exploring this line of work is beyond the scope of this paper and is left for future work.

\FloatBarrier

\newpage

\FloatBarrier
\clearpage
\newpage
\section{All fine-tune accuracies and picked models}

Finally, we provide plots that summarize all the results of the conducted large-scale experiment in a single overview per pool. The plots highlight the range of test accuracies amongst all the fine-tuned models, as well as the returned top-1 models ($B=1$) for the three strategies -- task-agnostic, task-aware linear and task-aware $k$NN.

\FloatBarrier

\begin{figure}[!htb]
\centering\captionsetup{width=1.0\linewidth}
\includegraphics[width=1.0\textwidth]{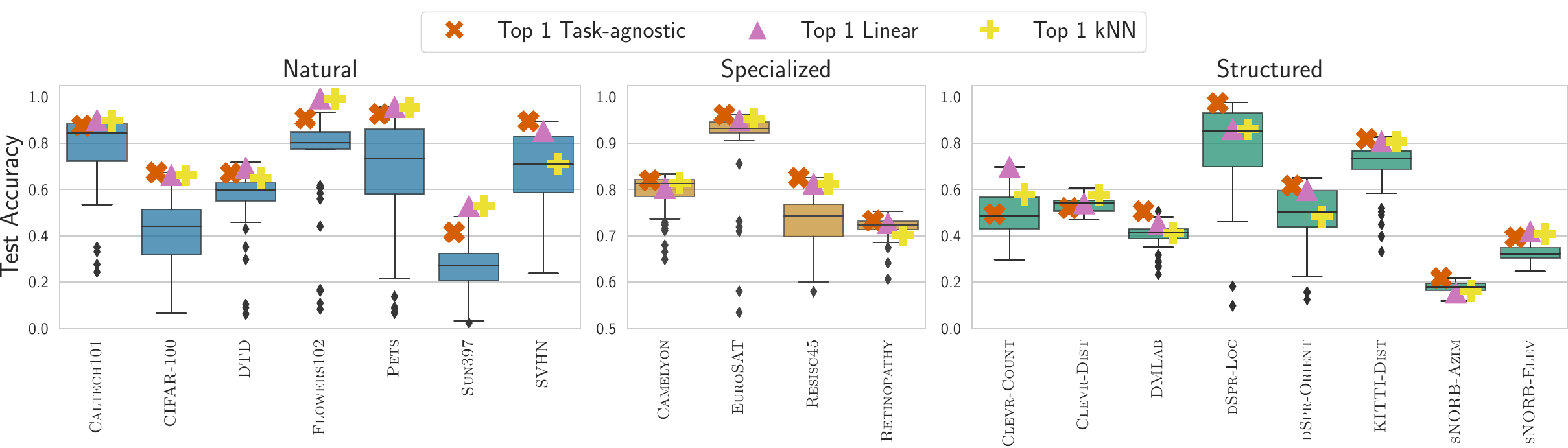}
\caption{Pool \all.}
\end{figure}

\begin{figure}[!htb]
\centering\captionsetup{width=1.0\linewidth}
\includegraphics[width=1.0\textwidth]{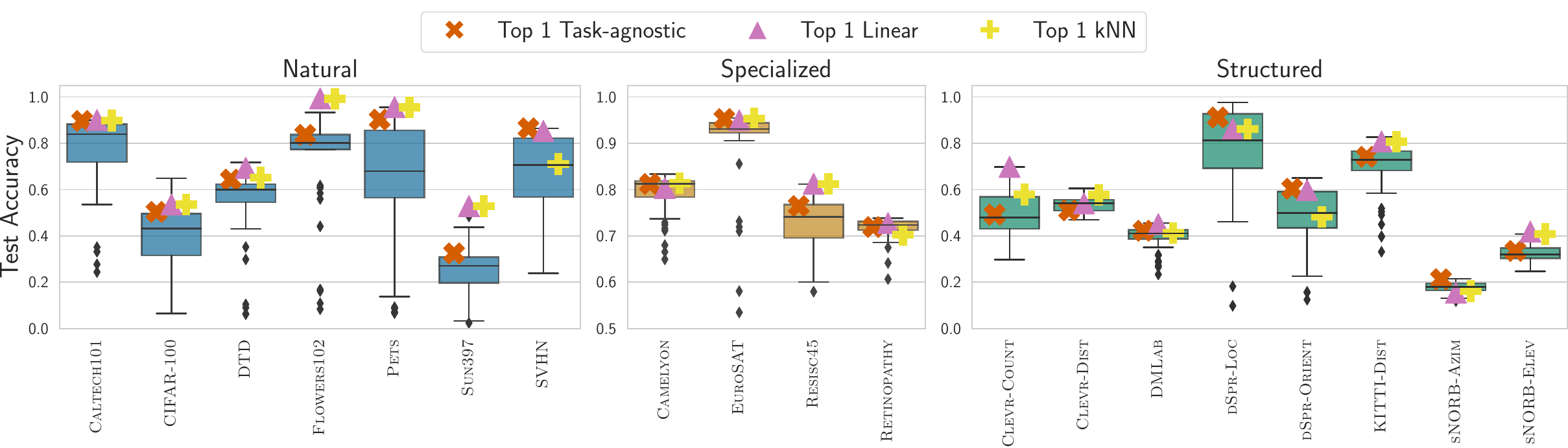}
\caption{Pool \restrdim.}
\end{figure}

\begin{figure}[!htb]
\centering\captionsetup{width=1.0\linewidth}
\includegraphics[width=1.0\textwidth]{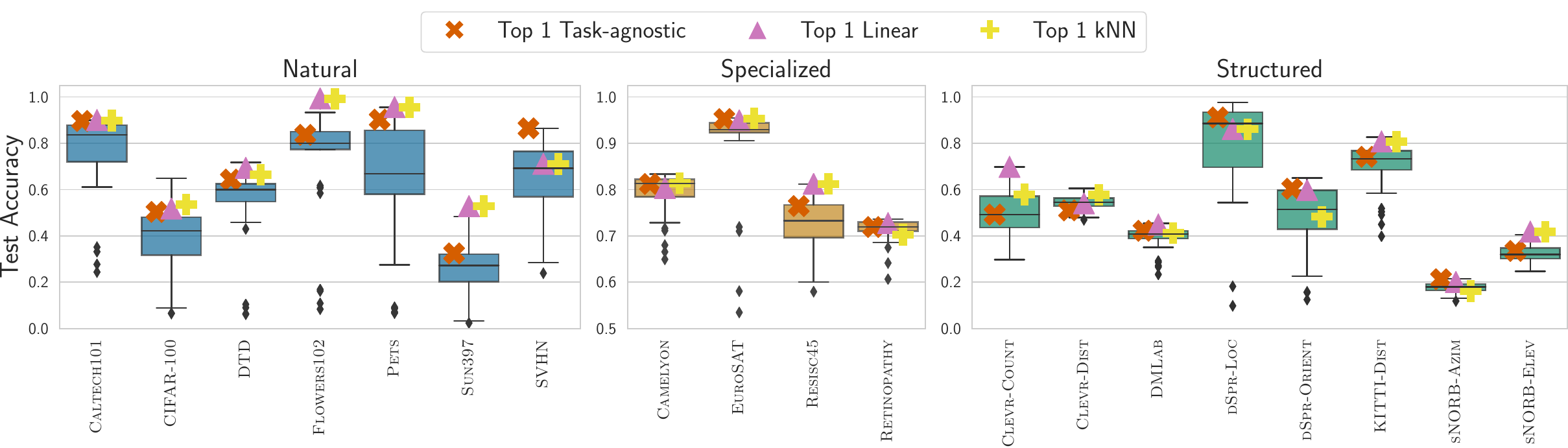}
\caption{Pool \maxresnet.}
\end{figure}

\newpage

\begin{figure}[!htb]
\centering\captionsetup{width=1.0\linewidth}
\includegraphics[width=1.0\textwidth]{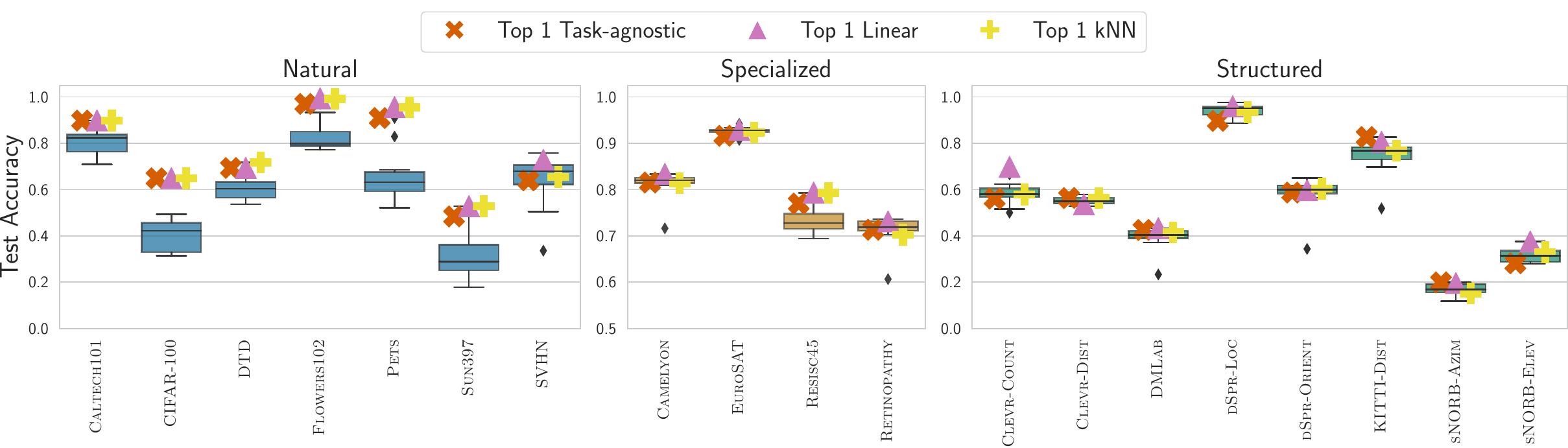}
\caption{Pool \jftslices.}
\end{figure}

\begin{figure}[!htb]
\centering\captionsetup{width=1.0\linewidth}
\includegraphics[width=1.0\textwidth]{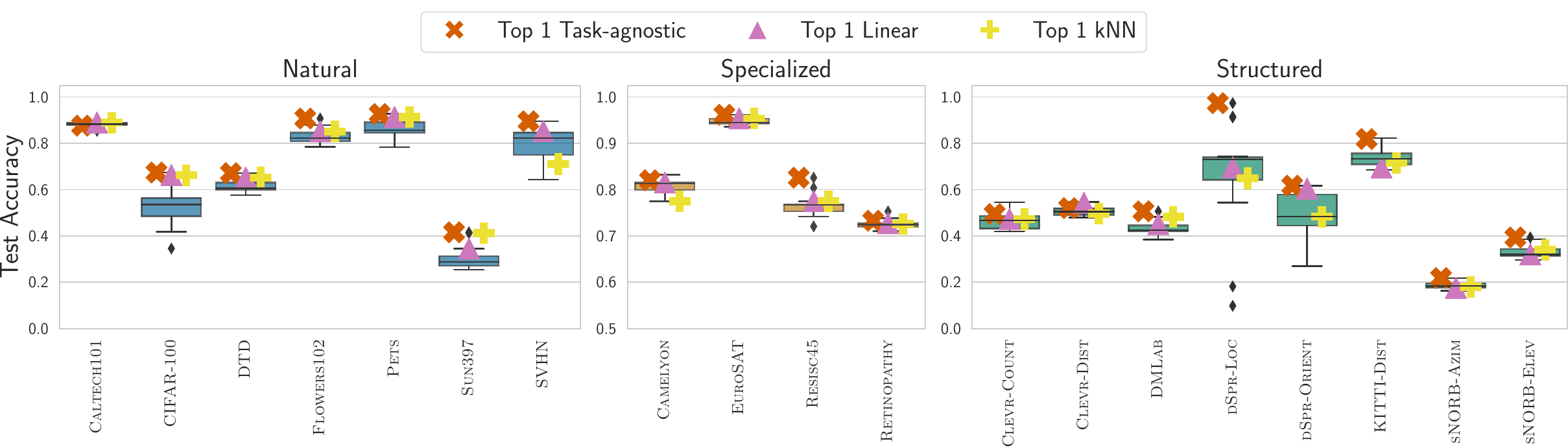}
\caption{Pool \imnetacc.}
\end{figure}

\FloatBarrier

\end{document}